\documentclass[10pt]{amsart} 
\usepackage{latexsym} 
\usepackage{amssymb} 
\usepackage{amscd}
\usepackage{epsf} 
\usepackage{graphicx}
\usepackage{verbatim}
\usepackage{longtable}
\usepackage{setspace}

\begin{document}

\newtheorem{Thm}{Theorem}[section] \newtheorem{TitleThm}[Thm]{} 
\newtheorem*{TitleThm1}{Theorem 1} 
\newtheorem*{TitleThm2}{Corollary 2} 
\newtheorem*{TitleThm3}{Theorem 3} 
\newtheorem*{TitleThm4}{Theorem 4} 
\newtheorem*{TitleThm5}{Theorem 5} 
\newtheorem*{TitleThm6}{Theorem 6} 
\newtheorem*{TitleThm7}{Corollary 7} 
\newtheorem*{TitleThm8}{Theorem 8} 
\newtheorem*{TitleThm9}{Theorem 9} 
\newtheorem*{TitleThm10}{Theorem 10}

\newtheorem{Corollary}[Thm]{Corollary} 
\newtheorem{Proposition}[Thm]{Proposition} 
\newtheorem{Lemma}[Thm]{Lemma} 
\newtheorem{Conjecture}[Thm]{Conjecture} 
\theoremstyle{definition} 
\newtheorem{Definition}[Thm]{Definition} 
\theoremstyle{definition} 
\newtheorem{Example}[Thm]{Example} 
\newtheorem{TitleExample}[Thm]{} 
\newtheorem{Remark}[Thm]{Remark} 
\newtheorem{SimpRemark}{Remark} 
\renewcommand{\theSimpRemark}{} 
\numberwithin{equation}{section}
\newcommand{\C}{{\mathbb C}}
\newcommand{\R}{{\mathbb R}} 
\newcommand{\Z}{{\mathbb Z}}
\newcommand{\bbP}{{\mathbb P}}

\newcommand{\flushpar}{\par \noindent}

\newcommand{\codim}{{\rm codim}\,} 
\newcommand{\proj}{{\rm proj}} 
\newcommand{\coker}{{\rm coker}\,}
\newcommand{\vol}{{\rm vol}\,}
\newcommand{\emb}{{\rm Emb}\,} 
\newcommand{\half}{\frac12} 
\newcommand{\wt}{{\rm wt}\,} 

\newcommand{\supp}{{\rm supp}\,}
\newcommand{\grad}{{\rm grad}\,}
\newcommand{\graph}{{\rm graph}\,}
\newcommand{\sing}{{\rm sing}\,}
\newcommand{\bdyM}{\partial M}
\newcommand{\Diff}{{\rm Diff}\,}
\newcommand{\intr}{{\rm int}\,}
\newcommand{\ptgG}{\partial \Gamma}
\newcommand{\Cinf}{\rm C^{\infty}} 
\newcommand{\ol}{\overline}
\newcommand{\ul}{\underline}
\newcommand{\barbdyM}{\overline{\partial M}}
\newcommand{\cirsp}[1]{\stackrel{\circ}{#1}}
\newcommand{\cirX}{\stackrel{\circ}{X}}
\newcommand{\ind}{{\rm ind}} 
\newcommand{\pd}[2]{\dfrac{\partial#1}{\partial#2}} 
\newcommand{\vnG}[1]{\mbox{vol}_{n+1}(\Gamma_t( #1 ))} 
\newcommand{\vnpG}[1]{\mbox{vol}_n(\partial \Gamma_t( #1 ))}

\def \ba {\mathbf {a}} 
\def \bb {\mathbf {b}} 
\def \bc {\mathbf {c}} 
\def \bD {\mathbf {D}}
\def \bA {\mathbf {A}}
\def \be {\mathbf {e}}
\def \bh {\mathbf {h}} 
\def \bk {\mathbf {k}}
\def \bl {\mathbf {\ell}} 
\def \bm {\mathbf {m}} 
\def \bn {\mathbf {n}} 
\def \bs {\mathbf {s}}
\def \bt {\mathbf {t}} 
\def \bu {\mathbf {u}} 
\def \bv {\mathbf {v}} 
\def \bx {\mathbf {x}} 
\def \by {\mathbf {y}}
\def \bw {\mathbf {w}} 
\def \b1 {\mathbf {1}}

\def \bzero {\boldsymbol {0}}
\def \bga {\boldsymbol \alpha} 
\def \bgb {\boldsymbol \beta} 
\def \bgg {\boldsymbol \gamma}
\def \bgW {\boldsymbol \Omega}
\def \bgD {\boldsymbol \Delta}

\def \itc {\text{\it c}} 
\def \iti {\text{\it i}} 
\def \itj {\text{\it j}} 
\def \itm {\text{\it m}} 
\def \itM {\text{\it M}}  
\def \ithn {\text{\it hn}} 
\def \itt {\text{\it t}}

\def \cA {\mathcal{A}} 
\def \cB {\mathcal{B}} 
\def \cC {\mathcal{C}} 
\def \cD {\mathcal{D}} 
\def \cE {\mathcal{E}} 
\def \cF {\mathcal{F}} 
\def \cG {\mathcal{G}} 
\def \cH {\mathcal{H}} 
\def \cI {\mathcal{I}}
\def \cJ {\mathcal{J}}
\def \cK {\mathcal{K}} 
\def \cL {\mathcal{L}} 
\def \cM {\mathcal{M}} 
\def \cN {\mathcal{N}} 
\def \cO {\mathcal{O}} 
\def \cP {\mathcal{P}} 
\def \cQ {\mathcal{Q}} 
\def \cR {\mathcal{R}} 
\def \cS {\mathcal{S}} 
\def \cT {\mathcal{T}} 
\def \cU {\mathcal{U}} 
\def \cV {\mathcal{V}} 
\def \cW {\mathcal{W}} 
\def \cX {\mathcal{X}} 
\def \cY {\mathcal{Y}} 
\def \cZ {\mathcal{Z}}

\def \ga {\alpha} 
\def \gb {\beta} 
\def \gg {\gamma} 
\def \gd {\delta} 
\def \ge {\epsilon} 
\def \gevar {\varepsilon} 
\def \gk {\kappa} 
\def \gl {\lambda} 
\def \gs {\sigma} 
\def \gt {\tau} 
\def \gw {\omega} 
\def \gz {\zeta} 
\def \gG {\Gamma} 
\def \gD {\Delta} 
\def \gL {\Lambda} 
\def \gS {\Sigma} 
\def \gW {\Omega}
\def \frm {\mathfrak{m}}

\def \dim {{\rm dim}\,} 
\def \mod {{\rm mod}\;}


\title[Medial/Skeletal Linking Structures] {Modeling Multi-Object 
Configurations via Medial/Skeletal Linking Structures} 
\author[James Damon and Ellen Gasparovic ]{James Damon$^1$ and Ellen 
Gasparovic$^2$} 
\thanks{(1) Partially supported by the Simons Foundation grant 230298, 
the National Science Foundation grant DMS-1105470 and DARPA grant 
HR0011-09-1-0055. (2) This paper contains work from this author's Ph. D. 
dissertation at Univ. of North Carolina} 
\address{Dept. of Mathematics \\ 
University of North Carolina \\ 
Chapel Hill, NC 27599-3250 }
\email{jndamon@math.unc.edu}
\address{Dept. of Mathematics \\
 Union College \\
Schenectady, NY 12308
}
\email{gasparoe@union.edu}

\begin{abstract}  
We introduce a method for modeling a configuration of objects in 2D or 3D 
images using a mathematical \lq\lq skeletal linking structure\rq\rq 
which will simultaneously capture the individual shape features of the 
objects and their positional information relative to one another.  The 
objects may either have smooth boundaries and be disjoint 
from the others or share common portions of their boundaries with other 
objects in a piecewise smooth manner.  These structures include a special 
class of \lq\lq Blum medial linking structures\rq\rq, which are 
intrinsically associated to the configuration and build upon the Blum 
medial axes of the individual objects.  We give a classification of the 
properties of Blum linking structures for generic configurations.  \par 
The skeletal linking structures add increased flexibility for modeling 
configurations of objects by relaxing the Blum conditions and they extend 
in a minimal way the individual \lq\lq skeletal structures\rq\rq which 
have been previously used for modeling individual objects and capturing 
their geometric properties.  This allows for the mathematical methods 
introduced for single objects to be significantly extended to the entire 
configuration of objects.  These methods not only capture the internal 
shape structures of the individual objects but also the external structure 
of the neighboring regions of the objects.  \par 
In our subsequent second paper \cite{DG2} we use these structures to 
identify specific external regions which capture positional information 
about neighboring objects, and we develop numerical measures for 
closeness of portions of objects and their significance for the 
configuration.  This allows us to use the same mathematical structures to 
simultaneously analyze both the shape properties of the individual objects 
and positional properties of the configuration.  This provides a framework 
for analyzing the statistical properties of collections of similar 
configurations such as for applications to medical imaging.  
\end{abstract}


\maketitle
\vspace{3ex}
\vspace{3ex}
\section{Introduction}  
\label{S:sec0} \par
\begin{figure}[ht]
\begin{center} 
\includegraphics[width=4cm]{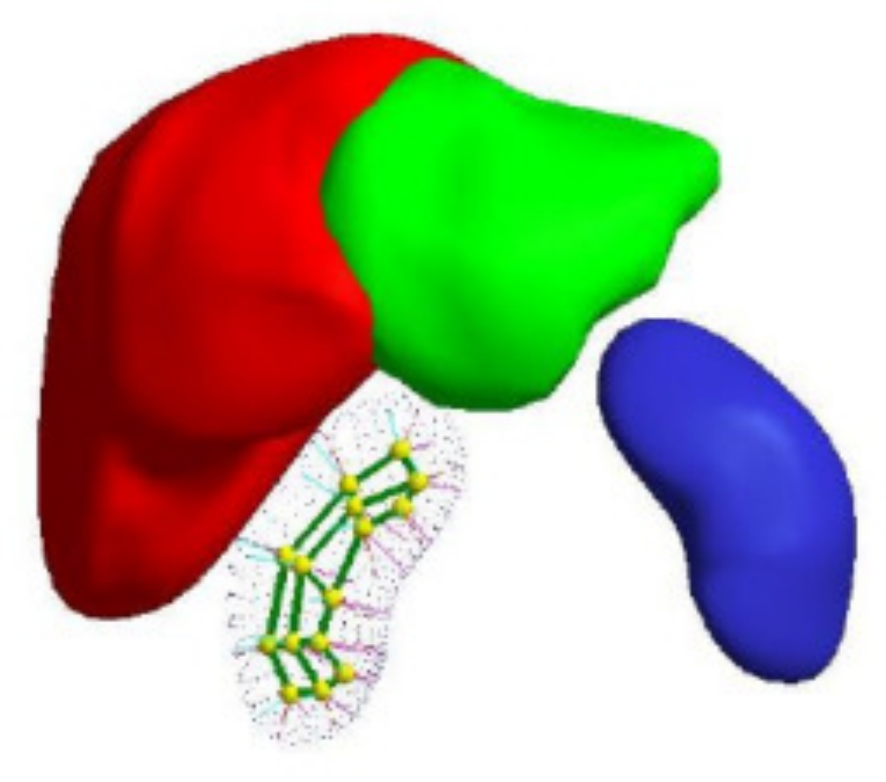}  \hspace{.25in} 
\includegraphics[width=5cm]{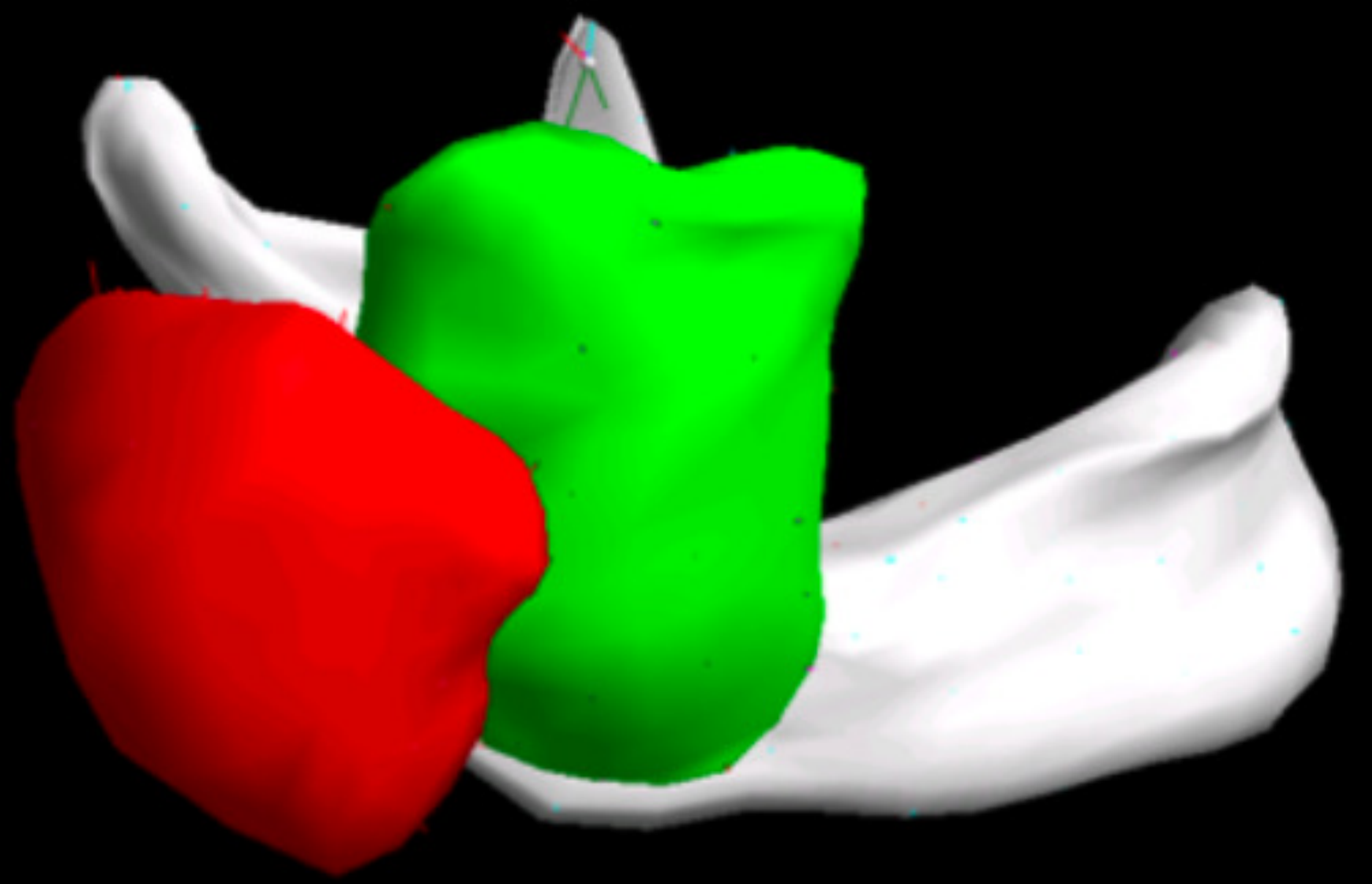}
\end{center}
\hspace{0.4in} (a) \hspace{1.8in} (b) 
 
\caption{\label{fig.1b} Examples of 3-dimensional medical images 
(obtained by MIDAG at UNC Chapel Hill) of a collection of physiological 
objects which can be modeled by a multi-region configuration.  
a)  Prostate, bladder and rectum in pelvic region \cite{CP} and b) 
mandible, masseter muscle, and parotid gland in throat region (modeled 
with a skeletal linking structure).}  
\end{figure}

Given a collection of objects, a basic question is how we may 
simultaneously model both the shapes of the individual objects and their 
positional information relative to one  another.  Motivating examples are 
provided by 2D and 3D medical images, in which we encounter collections 
of objects which might be organs, glands, arteries, bones, etc (see e.g. 
Figure \ref{fig.1b}).  A number of researchers have already begun to 
recognize the importance of using the relative positions of objects in 
medical images to aid in analyzing physical features for diagnosis and 
treatment (see especially the work of Stephen Pizer and coworkers in 
MIDAG at UNC for both time series of a single patient and for populations 
of patients \cite{LPJ}, \cite{GSJ}, \cite{JSM}, \cite{JPR}, \cite{Jg}, and 
\cite{CP}).  \par 
In these papers, significant use is made of the mathematical methods 
which are applied when modeling a single object using a \lq\lq skeletal 
structure\rq\rq \cite{D1}, \cite{D3}.  Such a structure generally consists 
of a \lq\lq skeletal set\rq\rq, which is a stratified set within the object, 
together with a multi-valued \lq\lq radial vector field\rq\rq\, from 
points on the set to the boundary of the object.  These form a flexible 
class of structures obtained by relaxing conditions on Blum medial axes, 
and these allow one to overcome various shortcomings of the Blum medial 
axis for modeling purposes.  They have played an important role by 
providing mathematical tools and numerical criteria for fitting discrete 
deformable templates as models for the individual objects in 3D medial 
images.  These discrete templates, called \lq\lq M-reps\rq\rq\ or in their 
more general forms \lq\lq S-reps\rq\rq, are discrete versions of skeletal 
structures where the skeletal set is a polyhedral surface with boundary 
edge, which is topologically a $2$-disk (versus the Blum medial axis 
which would have singularities), and with vectors defined at the vertices.  
In turn the fitting of the discrete deformable templates to individual 
objects in the medical images combine the numerical criteria with 
statistical priors formed from training sets of images.  In carrying out 
the stages of fitting of the models to the objects in the images, the 
mathematics of skeletal structures is used to guarantee nonsingularity of 
the models, for the interpolation of the discrete models to yield smooth 
surfaces, and for ensuring the regularity of resulting object model 
boundaries.  \par  
This approach has been successfully used in numerous cases for modeling 
objects in medical images, including e.g. the hippocampus, brain 
ventricles, corpus callosum, kidney, liver, and pelvic area involving the 
prostate, bladder and rectum.  For example, many references to papers 
that use M-reps and S-reps for the analysis of medical images of single 
objects are given in Chapters 1, 8 and 9 in the book edited and partially 
written by S. Pizer and K. Siddiqi, see reference [PS].  \par 
For certain collection of objects they have have been combined with user 
chosen, rather adhoc ways, to capture the relative positions between 
certain objects.  While considerable work has been devoted to the 
application to individual objects, there has not been developed an 
effective mathematical approach for an entire collection.  It is the goal of 
this paper to describe a mathematical framework for modeling an entire 
configuration of objects in 2D or 3D, which builds upon the success 
obtained for individual objects.
\par
\begin{figure}[ht] 
\includegraphics[width=6cm]{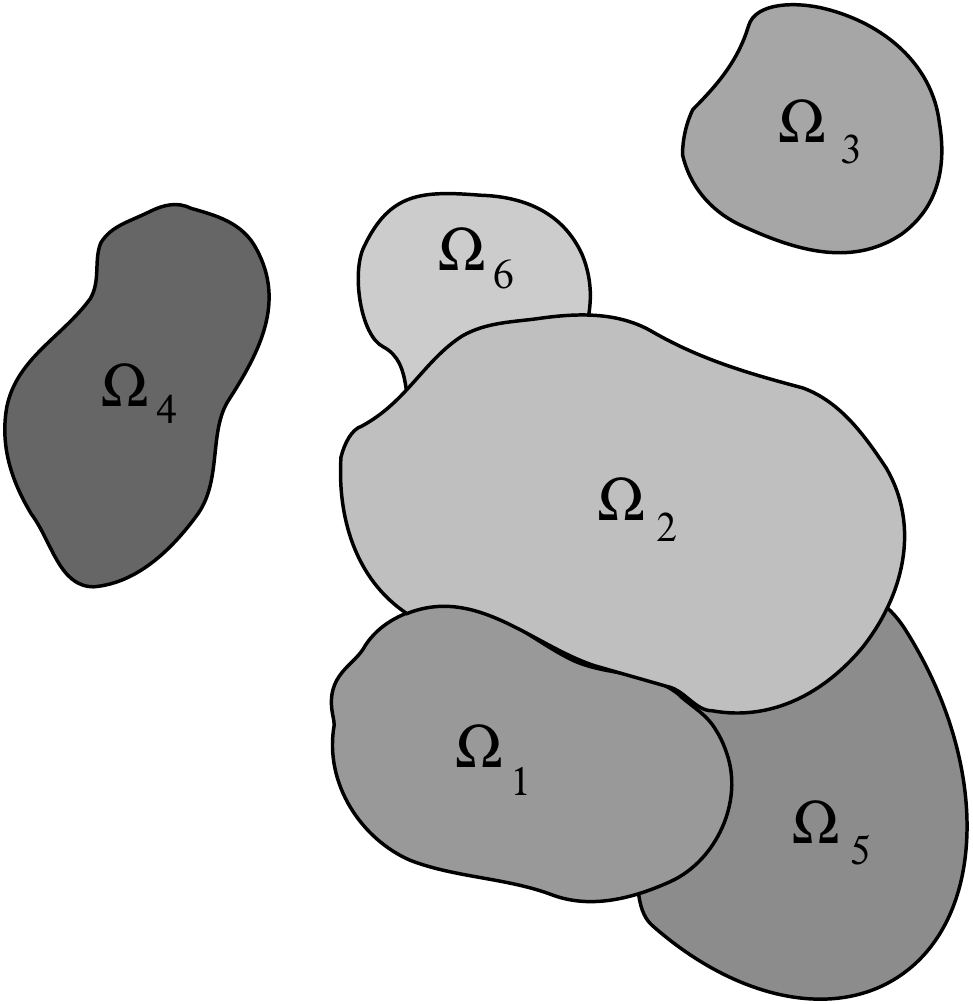} 
\caption{\label{fig.1a} Multi-region configuration in $\R^2$.} 
\end{figure}

 
We will concentrate on configurations of objects in 2D and 3D images.  
These can be modeled by a collection of distinct compact regions 
$\{ \gW_i\}$ in $\R^2$ or $\R^3$ with piecewise smooth generic 
boundaries $\cB_i$, where each pair of regions are either disjoint or only 
meet along portions of their boundaries in specified generic ways  (see 
e.g. Figure \ref{fig.1a}).  The mathematical results apply more generally to 
regions in $\R^n$ (see \cite{DG}) even though we concentrate here on 
$\R^2$ and $\R^3$. \par 
The geometric properties of the configuration are determined by both the 
shapes of the individual objects (regions) and their positions  in the 
overall configuration.  The \lq\lq shapes\rq\rq of the regions capture both 
the local and global geometry (and topology) of the regions.  The overall 
\lq\lq positional geometry\rq\rq of the configuration involves 
such information as: the measure of relative closeness of portions of 
regions, characterization of \lq\lq neighboring regions,\rq\rq and the 
\lq\lq relative significance\rq\rq of an individual region within the 
configuration.  \par 
Single numerical measures such as the Gromov-Hausdorff distance 
between objects in such configurations measure overall differences 
without identifying specific feature variations responsible for the 
differences.  Also invariants that would be appropriate for a collection of 
points fail to incorporate how shape differences between objects 
contribute to differences in positional structure of such configurations.  
We will introduce for such configurations \lq\lq 
medial and skeletal linking structures,\rq\rq which allow us to 
simultaneously capture local and global shape properties of the individual 
objects and their \lq\lq positional geometry.\rq\rq   \par
\par
In doing so we have three goals:
\begin{enumerate}
\item  minimizing the redundancy of the geometric information provided 
by the structure while allowing existing mathematical methods for single 
regions to be extended to configurations;
\item  robustness of the structure\rq s properties under small 
perturbations of the object\rq s shapes and positions in the configuration; 
and 
\item obtaining quantitative measures from the structure for use in 
statistical analysis of populations of \lq\lq configurations of the same 
type\rq\rq.
\end{enumerate}

\par
\par
To limit the amount of redundancy, the linking structures extend in a 
minimal way the notion of a \lq\lq skeletal structure\rq\rq for a single 
compact region $\gW$ with smooth boundary $\cB$, developed by the first 
author \cite{D1} (or see \cite{D3} for regions in $\R^2$ and $\R^3$).  It 
consists of a pair $(M, U)$, where the \lq\lq skeletal set\rq\rq $M$ is a 
\lq\lq Whitney stratified set\rq\rq in 
the region and $U$ is a multivalued \lq\lq radial vector field\rq\rq 
defined on $M$.  Skeletal structures are a more flexible class of 
structures which relax several of the conditions satisfied by the Blum 
medial axis of a region with smooth boundary \cite{BN}  The Blum medial 
axis is a special type of skeletal structure (with $U$ consisting of the 
vectors from points of $M$ to the points of tangency).  The resulting 
skeletal structure still retains the mathematical tools for analyzing the 
shape and geometric properties of the associated region.
\par
The added linking structure consists of a multi-valued \lq\lq linking 
function\rq\rq $\ell_i$ defined on the skeletal set $M_i$ for each region 
$\gW_i$ and a refinement of the Whitney stratification of $M_i$ on which 
$\ell_i$ is stratawise smooth.  The linking functions $\ell_i$ together 
with the radial vector fields $U_i$ yield multivalued \lq\lq linking vector 
fields\rq\rq $L_i$, which satisfy certain linking conditions.  Even though 
the structures are defined on the skeletal sets within the regions, the 
linking vector fields allow us to capture geometric properties of the 
external region as well.  \par 
A special type of skeletal linking structure is a {\it Blum medial linking 
structure}, which builds upon the Blum medial axes of the individual 
regions.  We classify the properties of the Blum linking structures for 
generic configurations.  In obtaining this structure, we identify the 
regions which remain unlinked and classify their local generic structure 
of their singular boundaries (which is also the singular set of the convex 
hull of the configuration).  We do this by introducing the {\it  spherical 
axis}, which is the analog of the Blum medial axis but for directions in 
$\R^n$, instead of distances; and it is defined using the family of height 
functions on the region boundaries.  \par
Just as for skeletal structures for single regions, the skeletal linking 
structure is a more flexible structure obtained by relaxing a number of the 
conditions for the Blum linking structure.  This allows more flexibility in 
applying skeletal structures to model objects.  Then, as we have already 
explained, this flexibility is what allows simplified skeletal structures 
to be used as discrete deformable templates for modeling objects 
\cite{P}, overcoming the lack of $C^1$-stability of the Blum medial axis, 
and providing discrete models to which statistical analysis can be applied 
\cite{P2}.  Likewise the skeletal linking structures have analogous 
properties, exhibiting stability under $C^1$ perturbations and allowing 
discrete models for applying statistical analysis.  This motivates their 
potential usefulness for modeling and computer imaging questions for 
medicine and biology.  
\par 
A skeletal structure enables both the local, relative, and global 
geometric properties of individual objects and their boundaries to be 
computed from the \lq\lq medial geometry\rq\rq of the radial vector field 
on the skeletal structure \cite{D2} and \cite{D4} (or see \cite{D3} for 
regions in $\R^2$ and $\R^3$).  This includes conditions ensuring the 
nonsingularity of a \lq\lq radial flow\rq\rq from the skeletal set to the 
boundary, providing a parametrization of the object interior by the level 
sets of the flow and implying the smoothness of the boundary \cite[Thm 
2.5]{D1}.  The medial/skeletal linking structure then extends these results 
to apply to the entire collection of objects, and includes the global 
geometry of the external region.  \par
\par 
In Section \ref{S:sec1}, we explain what we mean by multi-object 
configurations and how we mathematically model them using multi-region 
configurations.  Next, in Section \ref{S:sec2} we introduce 
medial/skeletal linking structures for multi-region configurations and 
give their basic properties.  In Section \ref{S:sec3}, we introduce two 
versions of a \lq\lq Blum linking structure\rq\rq for a general 
configuration.  We first give the generic properties of the Blum medial 
axis for a region with singular boundary having corners and edge curves (in 
the case of $\R^3$).  The Blum medial axis will now contain the singular 
points of the boundaries in its closure and we give an \lq\lq edge-corner 
normal form\rq\rq that the Blum medial axis will exhibit near such 
singular points.  We further give the local form of the stratification of the 
boundary by points associated to the singular points of the medial axis.  It 
is the generic interplay between these stratifications for adjoining 
regions and the linking medial axis that gives the generic properties of 
the linking structure.  \par 
For a generic multi-object configuration, which allows shared boundaries 
that have singular points, we establish the existence of a generic \lq\lq 
full Blum linking structure\rq\rq for the configuration (Theorem 
\ref{Thm3.6}); and later list the generic linking types for the 3D case in 
Section \ref{S:sec6}.  In the special case where all of the regions are 
disjoint with smooth generic boundaries, this directly yields a \lq\lq 
Blum medial linking structure\rq\rq (Theorem \ref{Thm3.5}).  This special 
case for disjoint regions was obtained in the thesis of the second author 
\cite{Ga}.  
In Section \ref{S:sec4} we explain how to modify for general 
configurations the resulting full Blum linking structure near the singular 
points of the boundaries to obtain a skeletal linking structure.  Lastly in  
Section \ref{S:sec7} we explain how the method involving M-reps and 
S-reps used as deformable templates for single regions can be expanded to 
obtain deformable templates for an entire configuration based on skeletal 
linking structures.
\par 
In a second paper \cite{DG2}, we will use the linking structure we have 
introduced to determine properties of the \lq\lq positional 
geometry\rq\rq of the configuration.  This will include: identifying 
distinguishing external regions capturing positional geometry; identifying 
neighboring regions via linking between these regions; introducing and 
computing numerical invariants of the positional geometry for measures 
of closeness and significance of regions in terms of volumetric 
measurements; computing volumetric invariants (which involve regions 
outside the 
configuration) as \lq\lq skeletal linking integrals\rq\rq; and combining 
these results to obtain a tiered graph structure which provides a 
hierarchical ordering of the regions.  This will provide for the comparison 
and statistical analysis of collections of objects in $\R^2$ and $\R^3$.  
\par 
The authors would like to thank Stephen Pizer for sharing with us his 
early work with his coworkers involving multiple objects in medical 
images.  This led us to seek a completely mathematical approach to these 
problems. 
\par

\section{Modeling Multi-Object Configurations in $\R^2$ and $\R^3$}  
\label{S:sec1}
\par
\subsection*{Local Models for Objects at Singular Points on Boundary}
We begin by defining what exactly we mean by a \lq\lq multi-object 
configuration.\rq\rq  We consider objects which have smooth boundaries 
if they are separated from the other objects.  However, if two objects 
meet along their boundaries then we allow two different situations, 
where objects either have flexible boundaries or are rigid.  The resulting 
possible configurations depend upon which combinations occur.  To model 
such meetings along boundaries, we first describe the form that the 
object\rq s boundaries take at the edges of such shared boundary regions.  
\par
\begin{figure}[ht]
\includegraphics[width=10cm]{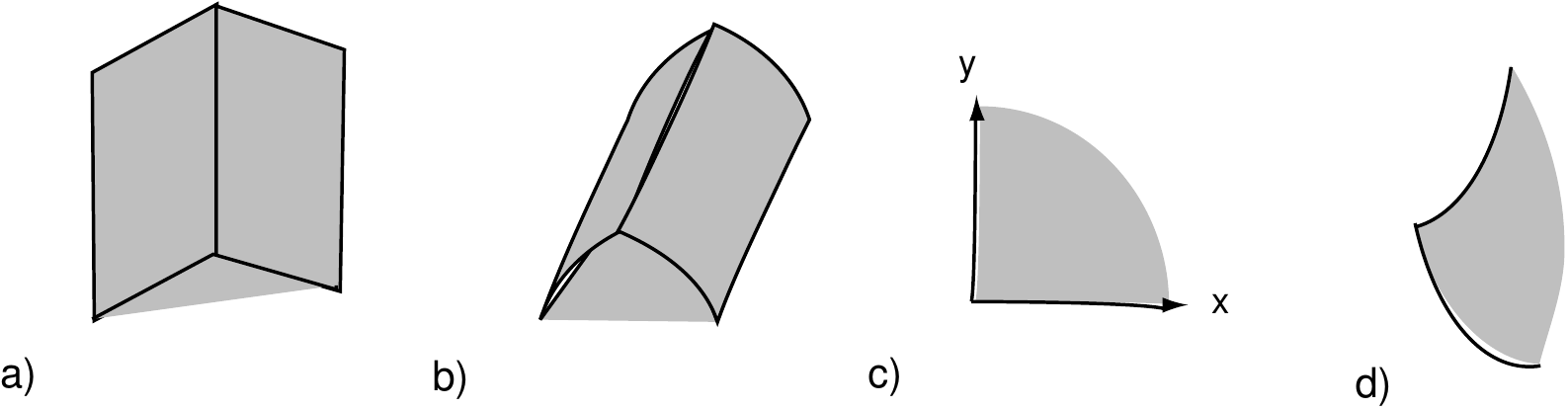}
\caption{\label{fig1.3a} Model for a (crease) edge in $\R^3$ a) and the 
general curved edge b); the model for a corner in $\R^2$ c) and the general 
curved corner d).}
\end{figure}
\begin{figure}[ht]
\includegraphics[width=8cm]{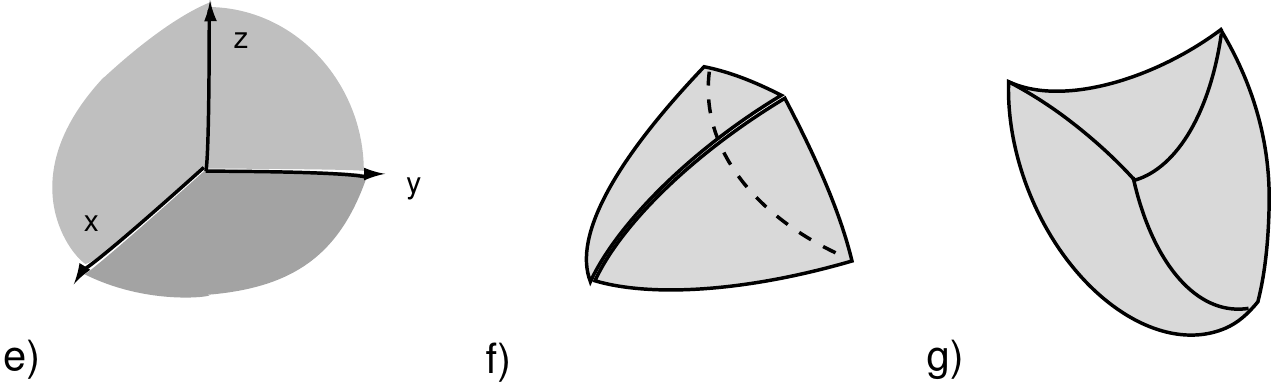}
\caption{\label{fig1.3b} Model for a convex corner in $\R^3$ e), and the 
corresponding curved convex corner f).  A concave corner g) may occur 
where three boundaries of regions which are pairwise mutually adjoined 
meet at a point.}
\end{figure}

 First, we model the individual objects by compact connected 2D or 3D 
regions $\gW$ with boundaries $\cB$ which are {\it smooth manifolds 
with boundaries and corners}.  We say that $\gW$ is a manifold with 
boundaries and corners if each point $x \in \cB$ is either a smooth point 
of the boundary or is modeled, via a diffeomorphic mapping of a 
neighborhood of the origin of $\R^2$, resp. $\R^3$, sending either a closed 
quadrant or half space of $\R^2$, resp. a closed octant, 
quarter-space, or half-space of $\R^3$ to a neighborhood of $x$ in $\gW$, 
with the boundaries of these regions in $\R^2$, resp. $\R^3$, mapping to 
the boundary $\cB$, see e.g. Figures~\ref{fig1.3a} and \ref{fig1.3b}.  
\par
\par
\par
\par
Then, the boundary $\cB$ is stratified by the corner points, the edge 
curves (for $\R^3$), and the smooth boundary regions (referred to as 
regular points of the boundary).  The union of the corner points and edge 
curves (in $\R^3$) form the singular points of the boundary.
\par
\begin{figure}[ht]
\begin{center} 
\includegraphics[width=10cm]{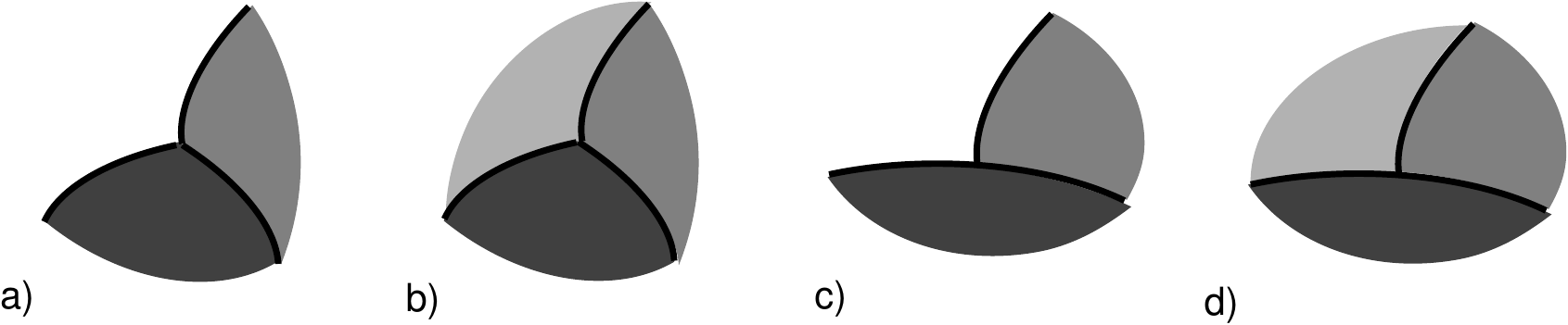} 
\end{center} 
\caption{\label{fig.2c} Generic local forms for adjoining regions in 
$\R^2$:  type $P_2$ consisting of a) two flexible regions meeting and b) 
three flexible regions meeting; or type $Q_2$ consisting of c) a flexible 
region meeting a rigid region and d) two flexible regions meeting a rigid 
region (with the darker rigid region below having the smooth boundary).}  
\end{figure}
\begin{figure}[ht]
\begin{center} 
\includegraphics[width=10cm]{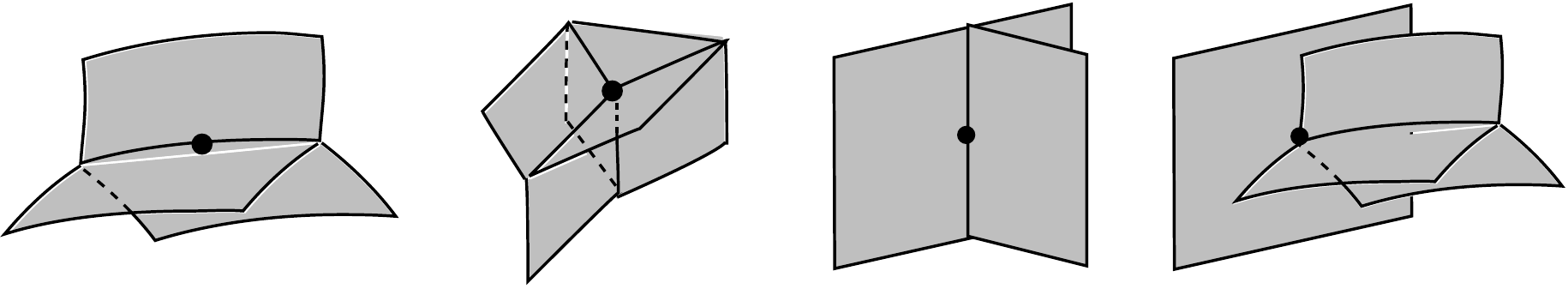} 
\end{center} 
\hspace{0.4in}(a) \hspace{0.8in} (b) \hspace {0.6in} (c) \hspace{0.8in} (d)
\caption{\label{fig.2b} Generic local forms for the boundaries where 
adjoining regions in 
$\R^3$ meet:  a) $P_2$ and b) $P_3$ represent the boundaries of several 
flexible regions meeting, including possibly the exterior region. Next, c) 
$Q_1$; and d) $Q_2$ represent the boundaries of flexible region(s) meeting 
a rigid region (which is to the left and has the smooth boundary).  The 
actual boundaries are diffeomorphic images of these models so the 
surfaces are in general curved and not planar. }  
\end{figure}

Second, we describe how we model objects sharing common boundary 
regions.  If two such regions intersect it will only be on their boundaries, 
and to describe the common boundary regions, we consider two cases 
depending on whether the regions are all flexible, or one of them is rigid.  
In the first case, for 3D objects, we model the edges of regions of common 
boundary regions by either a) $P_2$; b) $P_3$ in Figure \ref{fig.2b}. In 
these figures, one of the regions in the figures may denote the external 
region complementary to the objects.  If one of the regions is rigid then 
instead the regions meet as in c) $Q_1$ or d) $Q_2$ in Figure \ref{fig.2b}, 
where the region with smooth boundary is the rigid one, and again one of 
the other regions may denote the external region complementary to the 
objects.  For 2D objects, the local models for meeting are more simply 
given by either points of type $P_2$ in a) and b) in  Figure \ref{fig.2c} or 
type $Q_2$ in c) and d) in the same figure.  We refer to a configuration of 
regions satisfying the above two conditions regarding their boundary 
structure and their common boundary regions as satisfying the {\it 
combined boundary intersection condition}.  \par
\par
\par
\begin{Remark}
\label{Rem.2.1} 
Physically such local configurations of type a) or b) in Figures \ref{fig.2c} 
or \ref{fig.2b} are generic when objects with flexible boundaries have 
physical contact.  For example, these are typical of the singularities 
occurring where multiple soap bubbles are joined.  In general Maxwell\rq s 
Principle from physics implies that for boundaries of \lq\lq conflicting 
regions\rq\rq\, the singular boundaries have such specific forms 
(determined by a singularity theoretic analysis of the energy potential 
functions).  For c) and d), the region below the curve in Figure \ref{fig.2c} 
or to the left of the plane in Figure \ref{fig.2b} would represent a rigid 
region with which one or more flexible regions have contact.  
\end{Remark}
\par
We are going to model a multi-object configuration mathematically with a 
multi-region configuration, which we define next. 
\begin{Definition}
\label{Def.1.1}
A {\em multi-region configuration} consists of a collection of compact 
regions $\gW_i \subset \R^2$ or $\R^3$, $i=1,...,m$, with smooth 
boundaries and corners $\cB_i$ and satisfying the combined boundary 
intersection condition.  
\end{Definition}
\par
Then, a multi-region configuration will encode as data : i) the topological 
structure of each region; ii) the regions which share common boundaries  
and their individual types (flexible or rigid); and iii) the topological 
structure of the shared boundary regions between two regions.  Different 
regions having the same data will be viewed as having the same 
configuration type, and changes in the type will be viewed as resulting 
from a transition of types, see e.g. \S \ref{S:sec4}.  \par 
The configuration type can change due to: regions that were disjoint 
coming in contact, or the reverse; boundary regions of contact undergoing 
topological changes; regions undergoing changes in their topological type 
(e.g. a hole being created or destroyed); or more subtle positional changes.  
These are modeled by transitions of configuration type.  The generic 
transitions are the most common ones, but we will not attempt to 
classify or analyze them here.  If we wish to study variations and 
modifications within a single configuration type, then these can be most 
effectively described via an embedding mapping of a configuration which 
we consider next.  
\par
\subsection*{The Space of Equivalent Configurations via Mappings of a 
Model}
\par
We introduce a {\it model configuration} $\bgD$ for $\bgW$ in $\R^{n+1}$, 
$n=1,2$ which is a configuration of multi-regions $\{\gD_i\}$ satisfying 
Definition \ref{Def.1.1}.  A {\it 
configuration of the same type} as $\bgW$ will be obtained by an 
embedding $\Phi : \bgD \to \R^{n+1}$.  
This means that $\Phi : \cup_i \gD_i \to \R^{n+1}$ extends to a smooth 
embedding in some neighborhood, and it restricts to diffeomorphisms 
$\gD_i \simeq \gW_i$ for each $i$, with $\cB_i$ denoting the boundary of 
$\gW_i$.  Even though the configuration varies with $\Phi$ we still use 
the notation $\bgW$ for the resulting (varying) image configuration (with 
a specific $\Phi$ understood).  In particular, a configuration of model type 
$\bgD$ has all of the data of $\bgD$ built into it, so different regions 
$\gW_i$ and $\gW_j$ will meet along a shared boundary only when the 
corresponding $\gD_i$ and $\gD_j$ already do so and otherwise will not 
meet. \par 
The {\it space of configurations of type $\bgD$} is given by the infinite 
dimensional space of embeddings $\emb (\bgD, \R^{n+1})$.  Then, by {\it 
generic properties of a configuration} we will mean properties satisfied 
for an open dense set of embeddings in  $\emb (\bgD, \R^{n+1})$.  This 
means that for a configuration that exhibits generic properties, a 
sufficiently small perturbation will not destroy these properties, and if a 
configuration does not exhibit these properties, \lq\lq almost any 
arbitrarily small perturbation\rq\rq which is performed will ensure that 
it does.  The generic properties which we give are shown in \cite{DG} to 
satisfy this property.  \par
We say the regions $\gW_i$ and  $\gW_j$ are {\it adjoining regions} if 
$\cB_i \cap \cB_j \neq \emptyset$.  
Also, the  {\it external boundary region} of $\cB_i$ will denote that 
portion of the boundary which is shared by the external region. 
For example, in the multi-region configuration $\bgW$ in 
Figure~\ref{fig.1a} the regions $\gW_1$, $\gW_5$, $\gW_6$, are adjoined 
to $\gW_2$ (and  $\gW_1$ and $\gW_5$ with each other), and all of the 
regions have external boundary regions.  These features will persist for 
any configuration of the same type as this particular $\bgW$.
\par
\begin{Remark}
\label{Rem1.6}
In the special case of a configuration consisting of disjoint regions with 
smooth boundaries, each boundary is entirely external.
In such a case, we shall see that the geometric relations between the 
regions are entirely captured via \lq\lq linking behavior\rq\rq\, in the 
external region.  
\end{Remark}
\par
Because the external region extends indefinitely, we will frequently find 
for computational reasons that it is preferable to have the configuration 
contained in a \lq\lq bounded region\rq\rq.  By this we mean an ambient 
region $\tilde \gW$ so that $\gW_i \subset \tilde \gW$ for each $i$, then 
we say that $\bgW$ is a {\it configuration bounded by} $\tilde \gW$.  Such 
a $\tilde \gW$ might be a bounding box or disk or an intrinsic region 
containing the configuration.  Then we will also consider bounded 
configurations either given by an embedding $\tilde \Phi : \tilde \gD \to 
\R^2$ or $\R^3$, with $\tilde \Phi(\tilde \gD)$ denoting $\tilde \gW$, and 
$\Phi = \tilde \Phi | \bgD$; or we fix $\tilde \gW$ and consider 
embeddings $\Phi : \bgD \to \intr (\tilde \gW)$.  \par
\subsection*{Configurations Allowing Containment of Regions} 
\par
\begin{figure}[ht]
\includegraphics[width=7cm]{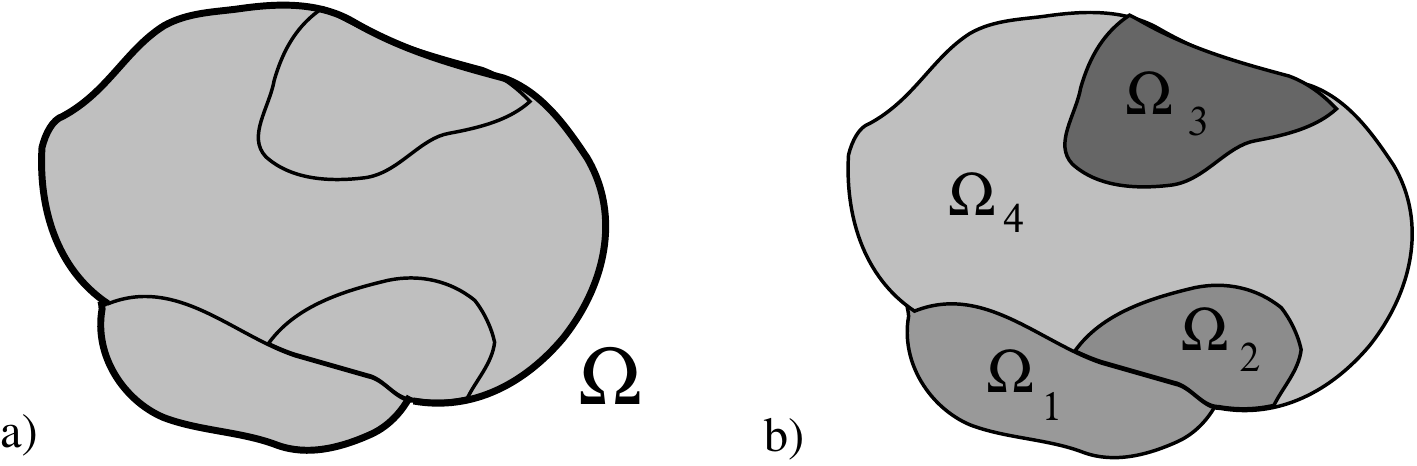}
\caption{\label{fig.1.5} a) is a configuration of regions contained in a 
region $\gW$.  It is equivalent to a multi-region configuration b) which is 
without inclusion.}
\end{figure}

In our definition of a multi-region configuration, we have explicitly 
excluded one region being contained in another.  However, given a 
configuration which allows this, we can easily identify such a 
configuration with the type we have already given.  To do so, if one region 
is contained in another  $\gW_i \subset \gW_j$, then we may represent 
$\gW_j$ as a union of two regions $\gW_i$ and the closure of $\gW_j 
\backslash (\intr(\gW_i) \cup (\cB_i \cap \cB_j))$, which we refer to as 
the {\it region complement} to $\gW_i$ in $\gW_j$.  By repeating this 
process a number of times we arrive at a representation of the 
configuration as a multi-region configuration in the sense of Definition 
\ref{Def.1.1}.  See Figure \ref{fig.1.5}.
\par
\par

\section{Skeletal Linking Structures for Multi-Region Configurations in 
$\R^2$ and $\R^3$}
\label{S:sec2}
\par
The skeletal linking structures for multi-object configurations will be 
constructed for the  multi-region configurations modeling them.  They 
will build upon the skeletal structures for individual regions.  We begin by 
recalling their basic definitions and simplest properties.  
\subsection*{Skeletal Structures for Single Regions}
\par
We begin by recalling \cite{D1} (or also see \cite{D3}) that a 
\lq\lq skeletal structure\rq\rq $(M, U)$ in $\R^{n+1}$ \cite[Def. 1.13]{D1}, 
consists of: a Whitney stratified set $M$ which satisfies the conditions 
for being a \lq\lq skeletal set\rq\rq\, \cite[Def. 1.2]{D1} and a 
multivalued \lq\lq radial vector field\rq\rq\, $U$ on $M$ which satisfies 
the conditions of \cite[Def. 1.5]{D1} and the 
\lq\lq local initial conditions\rq\rq \cite[Def. 1.7]{D1}.  
$M$ consists of smooth strata of dimension $n$, and the set of singular 
strata $M_{sing}$, with $\partial M$ denoting the singular strata where 
$M$ is locally a manifold with boundary (for which we use special \lq\lq 
boundary coordinates\rq\rq).  For images in $\R^2$ and $\R^3$, we are 
interested in the cases for $n=1, 2$; however, the general description is 
independent of dimension.  \par
\begin{figure}[ht]
\includegraphics[width=10cm]{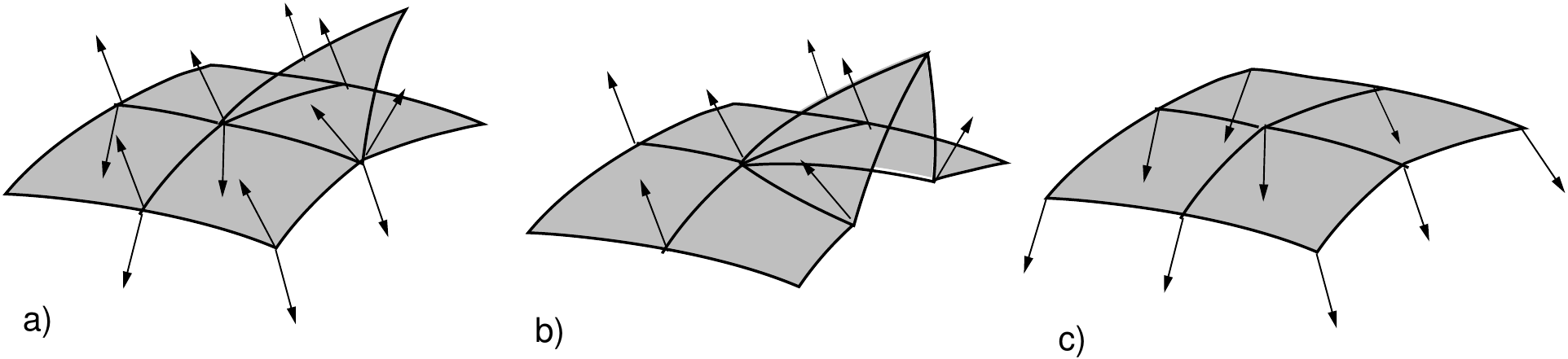}
\caption{\label{fig.10.4a} a) illustrates a neighborhood of a point in $M$  
and the multivalued vector fields .  b) and c) illustrate the two 
corresponding neighborhoods in $\tilde M$.}
\end{figure}

 
Without restating the conditions, we remark that these conditions allow 
each value of $U$ on a smooth point $x$ of $M$ to extend to a smooth 
vector field $U$ on a neighborhood of $x$, and various mathematical 
constructions on the smooth strata to be extended to the singular strata 
$M_{sing}$, see \cite[\S 2]{D1}.  Using the multivalued radial vector field 
$U$, we define a stratified set $\tilde M$, called the \lq\lq 
double of $M$\rq\rq\, and a finite-to-one stratified mapping $\pi : \tilde 
M \to M$, see \cite[\S 3]{D1}.  Points of $\tilde M$ consist of all pairs 
$\tilde x = (x, U)$ where $U$ is a value of the radial vector field at $x$, 
and $\pi (x, U) = x$ (see Figure~\ref{fig.10.4a}).  $\tilde M$ provides a 
mathematical method to keep track of both sides of each stratum of $M$.  
It also allows multivalued objects on $M$ to be viewed as single-valued 
objects on $\tilde M$.  
\par

\par 
Likewise we can keep track of the radial lines from each side of $M$ with 
the space $N_{+}= \tilde M \times [0, \infty)$, where for $\tilde x = (x, U) 
\in \tilde M$, the radial line (actually a half-line) is parametrized by $[0, 
\infty)$ by $c \mapsto c\cdot U$.  
Then, using $N_{+}$, we can define the \lq\lq radial flow\rq\rq.  In a 
neighborhood $W$ of a point $x_0 \in M$ with a smooth single-valued 
choice for $U$, we define a local representation of the radial flow by 
$\psi_t (x) = x + t\cdot U(x)$.  Together the local definitions yield a {\it 
global radial flow} as a mapping $\Psi: N_{+} \to \R^{n+1}$ defined for 
$\tilde x = (x, U)$ by $\Psi(\tilde x, t) = \psi_t (x)$.  
Beginning with a skeletal structure $(M, U)$, we can associate a \lq\lq 
region\rq\rq\, $\gW$ which is the image  $\Psi(N_{1})$, where $N_{1} = 
\tilde M \times [0, 1]$, and its \lq\lq boundary\rq\rq\, $\cB = \{ x + U(x) : 
x \in M\, \mbox{ all values of } U\}$.  
\par
\begin{figure}[ht]
\includegraphics[width=7cm]{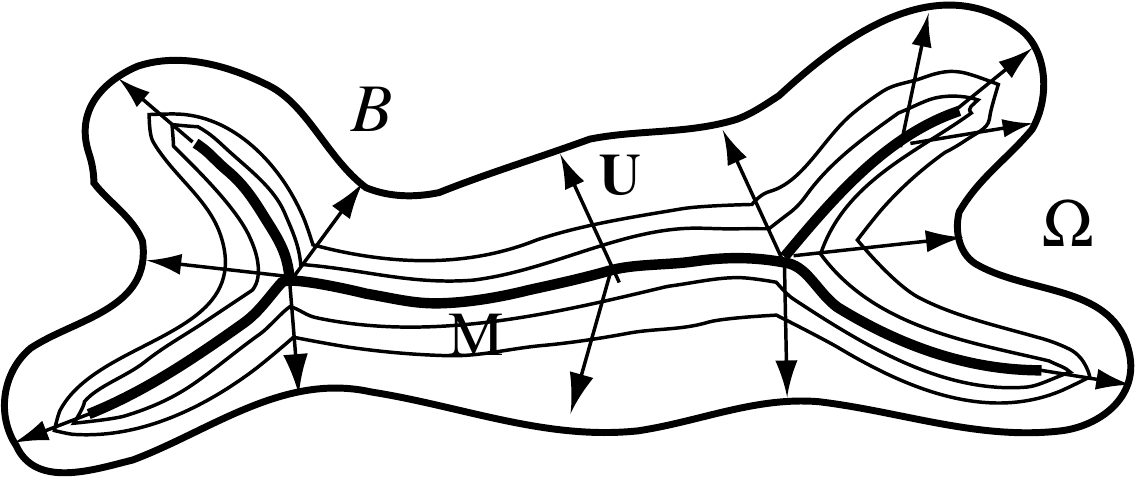}
\caption{\label{fig.II2c} Illustrating the radial flow from the skeletal set 
$M$, with the radial vector field $U$, flowing in the region $\gW$ to the 
boundary $\cB$, where the level sets are stratified sets forming a 
fibration of $\gW\backslash M$.}
\end{figure} 
\par
Provided certain curvature and compatibility conditions are satisfied, 
then by \cite[Thm 2.5]{D1} the radial flow defines a stratawise smooth 
diffeomorphism between $N_{1} \backslash \tilde M$ and $\gW \backslash 
M$; however the level sets of the flow $\cB_t$ are stratified rather than 
being smooth.  From the radial flow we define the radial map $\psi_1 (x) = 
x + U(x)$ from $\tilde M$ to $\cB$.  We may then relate the boundary and 
skeletal set via the radial flow and the radial map.  \par
A standard example we consider will be the Blum medial axis $M$ of a 
region $\gW$ with generic smooth boundary $\cB$ and its associated 
(multivalued) radial vector field $U$.  Then, the associated boundary $\cB$ 
will be the original boundary of the object.   \par
The {\it compatibility condition} involves the {\it compatibility 
$1$-form} $\eta_U = dr + \gw_U$, where $\| U\| = r$ is the {\it radial 
function} and $\gw_U(v) = v\cdot \bu$ where $U = r\cdot \bu$, with $\bu$ 
the multivalued unit vector field along $U$.  Then, $\eta_U$ is also 
multivalued with one value for each value of $U$. If $\eta_U$ vanishes on 
a neighborhood of $x$ in a smooth region of $M$, then $U(x)$ is orthogonal 
to the boundary $\cB$ at the point $\psi_1(x)$, and we refer to $U$ as 
being \lq\lq partially Blum\rq\rq at $x$.  Also, if $\eta_U$ vanishes on 
$M_{sing}$, then the boundary $\cB$ will be weakly $C^1$ on the image of 
the singular set.  However, if it does not, then the boundary will have 
corners and edges.  Hence, skeletal structures can also be used for regions 
with corners and edges.  \par 

\vspace{1ex}
\subsection*{Skeletal Linking Structures for Multi-Region Configurations} 
\hfill \par
We are ready to introduce skeletal linking structures for multi-region 
configurations.  These structures will accomplish multiple goals.  The 
most significant are the following.
\begin{itemize}
\item[i)]  Extend the skeletal structures for the individual regions in a 
minimal way to obtain a unified structure which also incorporates the 
positional information of the objects.  
\item[ii)] For generic configurations of disjoint regions with smooth 
boundaries, provide a Blum medial linking structure which incorporates 
the Blum medial axes of the individual objects. 
\item[iii)] For general multi-region configurations with common 
boundaries, provide for a modification of the resulting Blum structure to 
give a skeletal linking structure.
\item[iv)] Handle both unbounded and bounded multi-region configurations. 
\end{itemize} 
In Part II \cite{DG2} we shall also see that the skeletal linking structure 
has a second important function allowing us to answer various questions 
involving the \lq\lq positional geometry\rq\rq of the regions in the 
configuration.  
\begin{Remark}
Even though a configuration itself is bounded, the complement is 
unbounded.  The properties we shall give will be valid for the unbounded 
external region; however, for both computational and measurement 
reasons, it is desirable to consider the configuration as lying in a 
bounding region.  This is the \lq\lq bounded case\rq\rq.  There are a 
number of possibilities for such bounding regions: intrinsic bounding 
region, a bounding box, the convex hull, or a bounded region resulting from 
imposing a threshold, etc.  We shall consider further the relation of these 
with a skeletal linking structure modeling a given configuration at the end 
of this section.
\end{Remark}
 \par
We begin by giving versions of the definition for both the bounded and 
unbounded cases.  
\begin{Definition}
\label{Defmultlkgstr} A {\em skeletal linking structure for a multi-region 
configuration} $\{\gW_i\}$ in $\R^2$ or $\R^3$ consists of a triple $(M_i, 
U_i , 
\ell_i)$ for each region $\gW_i$ with the following properties.
\begin{itemize}
\item[S1)]
$(M_i, U_i)$ is a skeletal structure for $\gW_i$ for each $i$ 
with $U_i =  r_i\cdot \bu_i$ a scalar multiple of $\bu_i$ a (multivalued) 
unit vector field and $r_i$ the multivalued radial function on $M_i$.
\item[S2)] $\ell_i$ is a (multivalued) {\em linking function} defined on 
$M_i$ (excluding the strata $M_{i\, \infty}$, see L4 below), with one value 
for each value of $U_i$, for which the corresponding values satisfy $\ell_i 
\geq r_i$, and it yields a (multivalued) {\em linking vector field} $L_i = 
\ell_i\cdot \bu_i$ .
\item[S3)] The canonical stratification of $\tilde M_i$ has a stratified 
refinement $\cS_i$, which we refer to as the {\em labeled refinement}.
\end{itemize}
By $\cS_i$ being a \lq\lq labeled refinement\rq\rq of the stratification  
$\tilde M_i$ we mean it is a refinement in the usual sense of 
stratifications in that each stratum of $\tilde M_i$ is a union of strata of 
$\cS_i$; and they are labeled by the linking types which occur on the 
strata. \par 
In addition, they satisfy the following four {\em linking conditions}. 
\flushpar
{\it Conditions for a skeletal linking structure} \par
\begin{itemize}
\item[L1)]   $\ell_i$ and $L_i$ are continuous where defined on $M_i$ and 
are smooth on strata of $\cS_i$.

\item[L2)]  The \lq\lq linking flow\rq\rq (see (\ref{Eqn2.1}) below) 
obtained by extending the radial flow is nonsingular and for the strata 
$S_{i\, j}$ of $\cS_i$, the images of the linking flow are disjoint and each 
$W_{i\, j} = \{x + L_i (x): x \in S_{i\, j}\}$ is smooth.  

\item[L3)]  The strata $\{W_{i\, j}\}$ from the distinct regions either 
agree or are disjoint and together they form a stratified set $M_0$, which 
we shall refer to as the {\em (external) linking axis}.
\item[L4)]  There are strata $M_{i\, \infty} \subset \tilde M_i$ on which 
there is no linking so the linking function $\ell_i$ is undefined.  On the 
union of these strata $M_{\infty} = \cup_{i} M_{i\, \infty}$, the global 
radial flow restricted to $N_{+} | M_{\infty}$ is a diffeomorphism with 
image the complement of the image of the linking flow.  In the bounded 
case, with $\tilde \gW$ the enclosing region of the configuration, it is 
required that the boundary of $\tilde \gW$ is transverse to the 
stratification of $M_0$ and where the linking vector field extends beyond 
$\tilde \gW$, it is truncated at the boundary of $\tilde \gW$ (this 
includes $M_{\infty}$).  
\end{itemize}
\end{Definition}
\par
We denote the region on the boundary corresponding to $M_{\infty}$ by 
$\cB_{\infty}$ and that corresponding to $M_{i\, \infty}$ by 
$\cB_{i\, \infty}$. 
\par 
\begin{figure}[ht] 
\centerline{
\includegraphics[width=3.5cm]{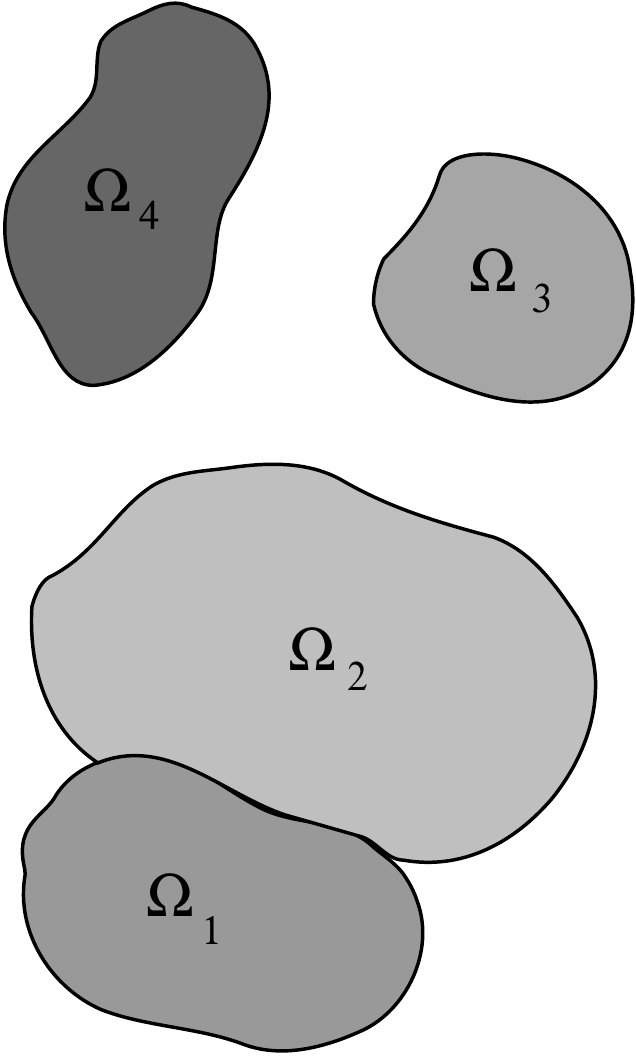}
\hspace{1.5cm}
\includegraphics[width=3.7cm]{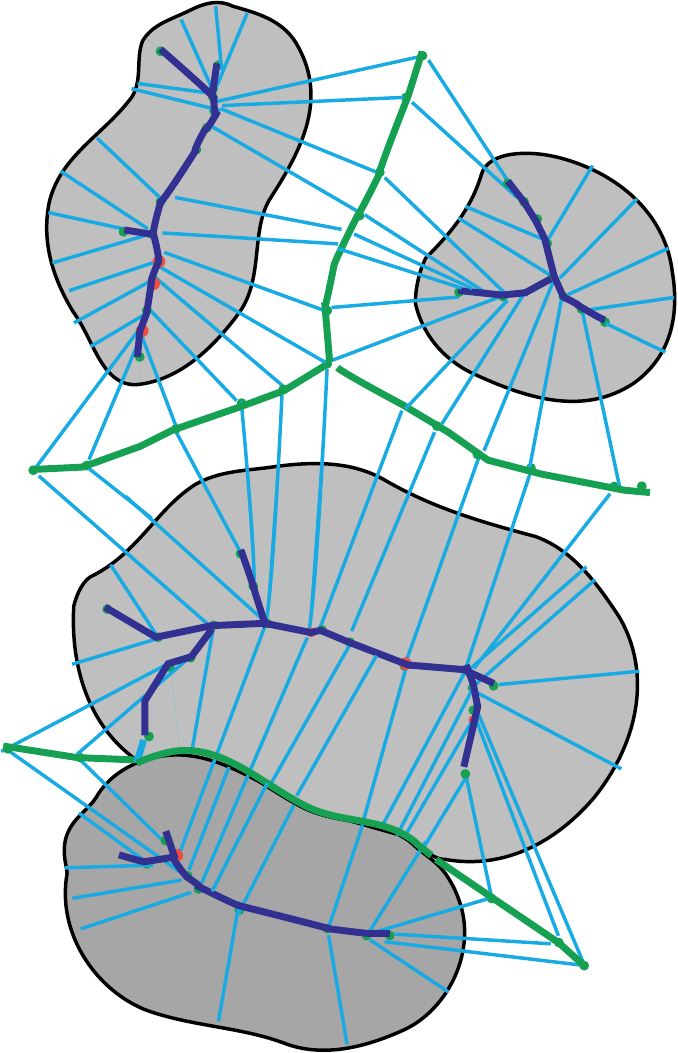}} 
\caption{\label{fig.5a} Multi-region configuration in $\R^2$ and (a portion 
of) the skeletal linking structure for the configuration.  Shown are the 
linking vector fields meeting on the (external) linking axis.  The linking 
flow moves along the lines from the medial axes to meet at the (external) 
linking  axis, which includes the portion of the adjoined boundaries of 
$\gW_1$ and $\gW_2$.} 
\end{figure} 
\par
\begin{Remark}
\label{Rem.2.1a}
By property L4), $M_{i\, \infty}$ does not exhibit any linking with any 
other region.  We will view it as either the {\em unlinked region} or 
alternately as being {\it linked to $\infty$}, where we may view the 
linking function as being $\infty$ on $M_{i\, \infty}$.  In the bounded case 
we modify the linking vector field so it is truncated at the boundary of 
$\tilde \gW$.  We can also introduce a \lq\lq linking vector field on 
$M_{\infty}$\rq\rq\, by extending the radial vector field until it meets 
the boundary of $\tilde \gW$ (see e.g. Figure \ref{fig.II4.2}).  
\end{Remark}
\par
For this definition, we must define the \lq\lq linking flow\rq\rq\,  which 
is an extension of the radial flow.  We define the {\it linking flow} from 
$M_i$ by 

\begin{align}
\label{Eqn2.1}
\gl_{i}(x,t) \, \, &= \,\, x \,+ \, \chi_i(x,t)\bu_i(x)  \qquad \text{where}  
\\
\chi_i(x,t) &= \left\{      
\begin{array}{lr}       
 2tr_i(x) &  \displaystyle 0 \leq t \leq \frac{1}{2}\\        2(1-t)r_i(x) + 
(2t-1)\ell_i(x) &  \displaystyle \frac{1}{2} \leq t \leq 1      
\end{array}    \right.  \, .   \notag
\end{align}
As with the radial flow it is actually defined from $\tilde M_i$ (or $\tilde 
M_i \backslash M_{\infty}$). The combined linking flows $\gl_i$ from all 
of the $M_i$ will be jointly referred to as the linking flow $\gl$.  For 
fixed $t$ we will frequently denote $\gl(\cdot, t)$ by $\gl_t$.
\vspace{1ex}
\flushpar
{\bf Convention: }  It is convenient to view the collection of objects for 
the linking structure as together forming a single object, so we will adopt 
notation for the entire collection.  This includes $M$ for the union of the 
$M_i$ for $i > 0$, and similarly for $\tilde M$.  On $M$ (or $\tilde M$) we 
have the radial vector field $U$ and radial function $r$ formed from the 
individual $U_i$ and $r_i$, the linking function $\ell$ and linking vector 
field $L$ formed from the individual $\ell_i$ and $L_i$; as well as the 
linking flow $\gl$ and $M_{\infty}$ already defined. 
\par
\vspace{1ex}
\par
We see that for $0 \leq t \leq \frac{1}{2}$ the flow is the radial flow at 
twice its speed; hence, the level sets of the linking flow, $\cB_{i\, t}$, 
for time $0 \leq t \leq \frac{1}{2}$ will be those of the radial flow.  For 
$\frac{1}{2} \leq t \leq 1$ the linking flow is from the boundary $\cB_i$ 
to the linking strata of the external medial linking axis.   \par 
By the {\it linking flow being nonsingular} we mean it is a piecewise 
smooth homeomorphism, which for each stratum $S_{i\, j} \subset \tilde 
M_i$, is smooth and nonsingular on $S_{i\, j} \times \left[ 0, 
\frac{1}{2}\right]$ and either: $S_{i\, j} \times \left[ \frac{1}{2}, 
1\right]$ is smooth and nonsingular if $S_{i\, j}$ is a stratum associated 
to strata in $\cB_{i\, 0}$.  For $S_{i\, j}$ not associated to strata in 
$\cB_{i\, 0}$, $\ell_i = r_i$ on $S_{i\, j}$, so the linking flow on $S_{i\, 
j} \times \left[ \frac{1}{2}, 1\right]$ is constant as a function of $t$.  
That the linking flow is nonsingular will follow from the analogue of the 
conditions given in \cite[\S 3]{D1} for the nonsingularity of the radial 
flow.  These will be given in Part II \cite{DG2}, when we use the linking 
flow to establish geometric properties of the configuration.
\par

\subsection*{Linking between Regions and between Skeletal Sets}
\par
A skeletal linking structure allows us to introduce the notion of linking of 
points in different (or the same) regions and of regions themselves being 
linked.    We say that two points $x \in \tilde M_i$ and $x^{\prime} \in 
\tilde M_j$ are {\em linked} if the linking flows satisfy $\gl_i(x, 1) = 
\gl_j(x^{\prime}, 1)$. This is equivalent to saying that for the values of 
the linking vector fields $L_i (x)$ and $L_j(x^{\prime})$, $x + L_i(x) = 
x^{\prime} + L_j (x^{\prime})$.  Then, by linking property $L3)$, the set of 
points in $\tilde M_i$ and $\tilde M_j$ which are linked consist of a union 
of strata of the stratifications $\cS_i$ and $\cS_j$.  Furthermore, if the 
linking flows on strata from $S_{i\, k} \subset \tilde M_i$ and $S_{j\, 
k^{\prime}} \subset 
\tilde M_j$ yield the same stratum $W \subset M_0$, then we refer to the 
strata as being linked via the linking stratum $W$.  Then, $\mu_{i\, j} = 
\gl_j(\cdot, 1)^{-1}\circ \gl_i(\cdot, 1)$ defines a diffeomorphic {\it 
linking correspondence} between $S_{i\, k}$ and $S_{j\, k^{\prime}}$.  
\par
\begin{figure}[ht] 
\centerline{\includegraphics[width=6cm]{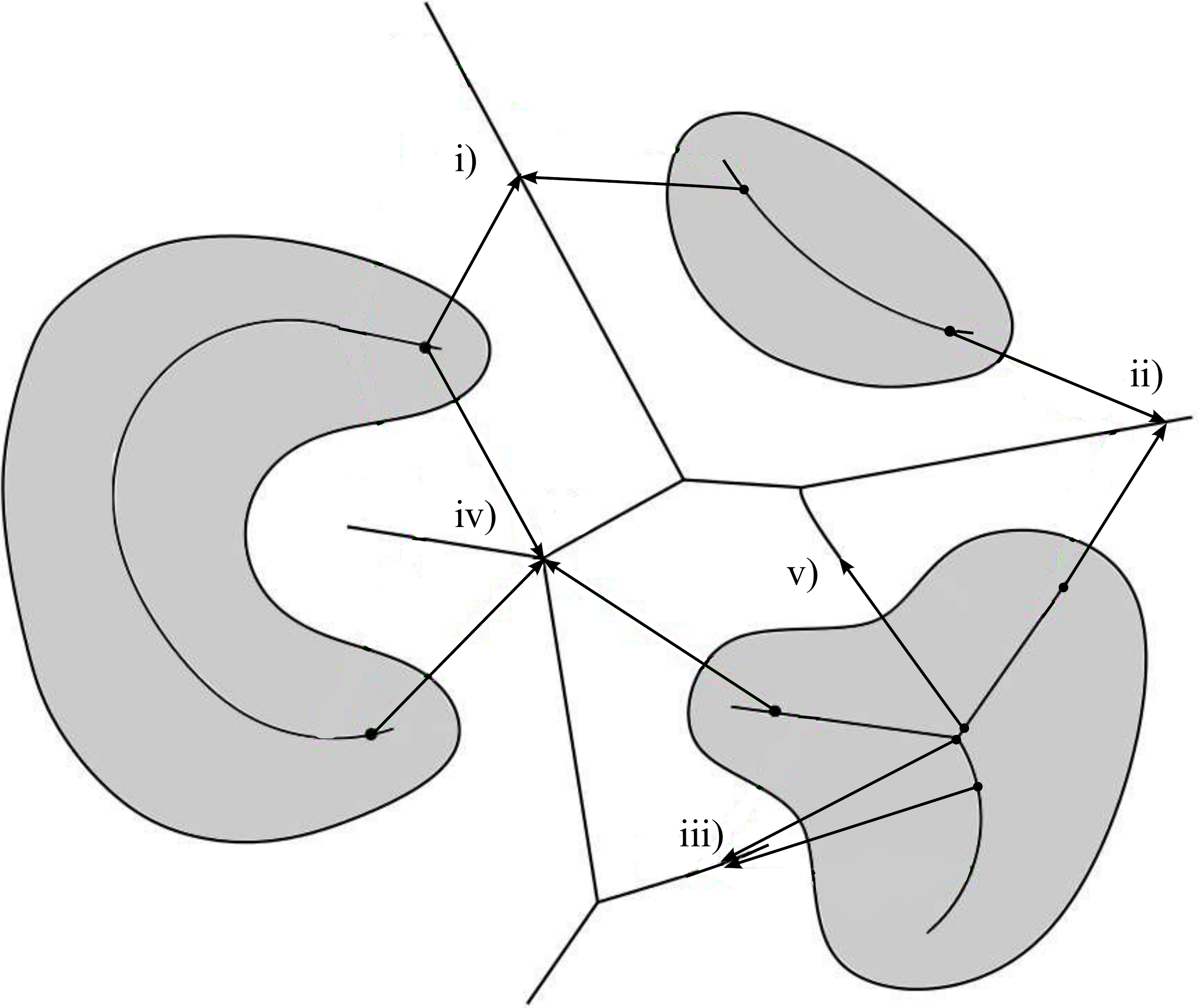}} 
\caption{\label{fig.5d} Types of linking for multi-region
configuration in $\R^2$: The numbers correspond to the generic Blum 
linking types listed in \S \ref{S:sec3}.  Also, i) and ii) illustrate linking 
between two objects; iii) and v) self-linking; and iv) partial linking.} 
\end{figure}

 
In Part II \cite{DG2} we will introduce a collection of regions which 
capture geometrically the linking relations between the different regions.  
For now we concentrate on understanding the types of linking that can 
occur.  
There are several possible different kinds of linking.  More than two 
points may be linked at a given point in $M_0$.  Of these more than one 
may be from the same region.  If all of the points are from a single region, 
then we call the linking {\it self-linking}, which occurs at indentations of 
regions.  If there is a mixture of self-linking and linking involving other 
regions then we refer to the linking as {\it partial linking}, see 
Figure~\ref{fig.5d}.  \par 
\par
\begin{Remark}
\label{Rem2.2}  
If $\gW_i$ and $\gW_j$ share a common boundary region, then certain 
strata in $M_i$ and $M_j$ are linked via points in this boundary region, 
and for those $x \in M_i$, $\ell_i(x) = r_i(x)$; see Figure~\ref{fig.5a}.  
\end{Remark}
\par
We next consider in more detail the bounded case.
\subsection*{Skeletal Linking Structures in the Bounded Case}
 \par
For the bounded case we suppose that the configuration lies in the interior 
of a \lq\lq bounding region\rq\rq $\tilde \gW$ so that the boundary 
$\partial \tilde \gW$ is transverse to: i) the strata of the external medial 
axis $M_0$, ii) the extension of the radial lines from $M_{\infty}$, and iii) 
the linking line segments.  If the boundary $\partial \tilde \gW$ has 
singular points, then for ii) and iii) by the lines being transverse to 
$\partial \tilde \gW$, we mean that at a singular point, the line is 
transverse to the limits of tangent spaces from the smooth points of 
$\partial \tilde \gW$.  For any  convex region $\tilde \gW$ with 
piecewise smooth boundary, the limiting tangent planes of $\partial \tilde 
\gW$ are supporting hyperplanes for $\tilde \gW$.  Hence any line in the 
tangent plane lies outside $\intr(\tilde \gW)$.  Thus, the radial lines from 
regions in the configuration will always meet the boundary transversely.  
\par 
We can alter the linking vector field either by truncating it where it 
meets the boundary $\partial \tilde \gW$ or defining it on $M_{\infty}$ 
and then refining the stratification so that on appropriate strata the 
linking vector field ends at $\partial \tilde \gW$.  It is now defined on all 
of $M_i$ for the individual regions $\gW_i$.  Because we are either 
reducing $\ell_i$, or defining $L_i$ on $M_{i\, \infty}$, the linking flow is 
still nonsingular.  

Thus, we have compact versions of the regions defined for the unbounded 
case.  There are a number of different possibilities for such bounding 
regions.  
\par
\begin{figure}[ht] 
\begin{center}
\includegraphics[width=4.0cm]{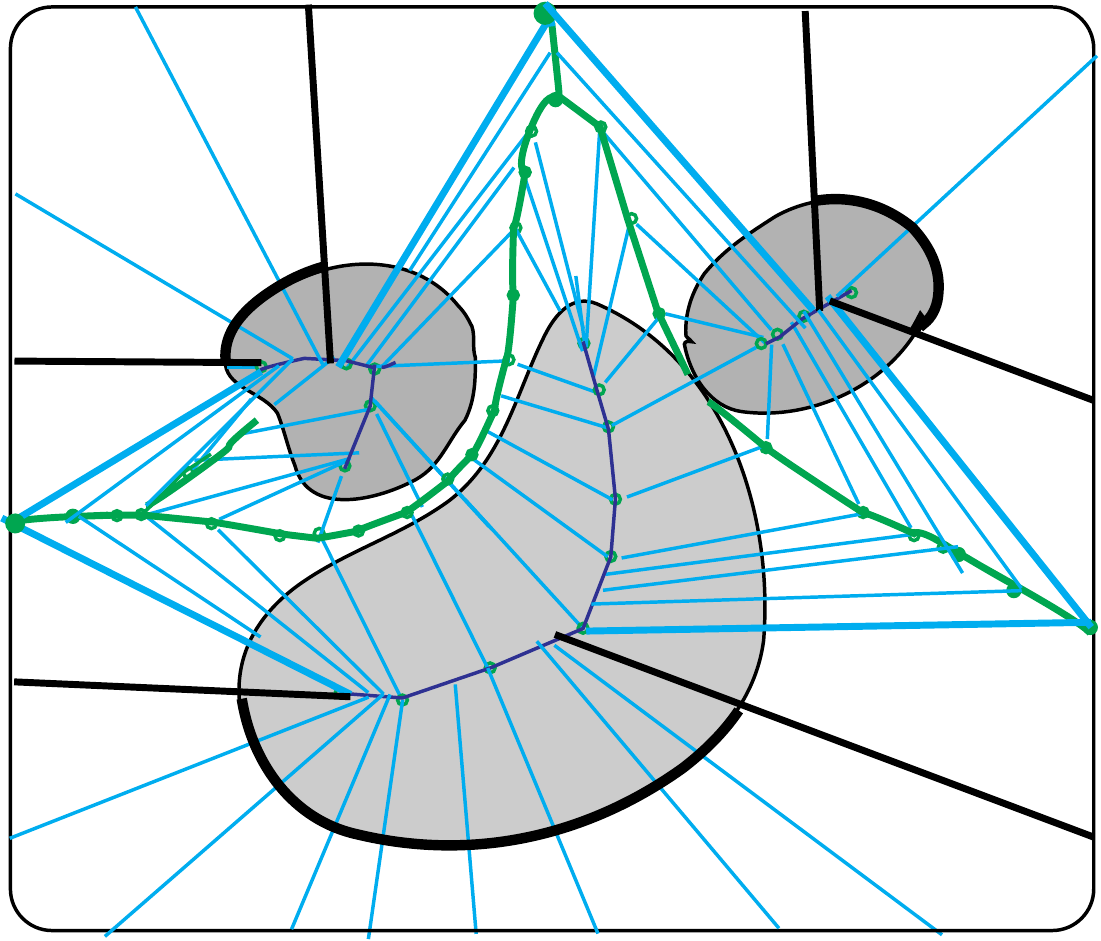} 
\hspace*{0.10cm}
\includegraphics[width=4.0cm]{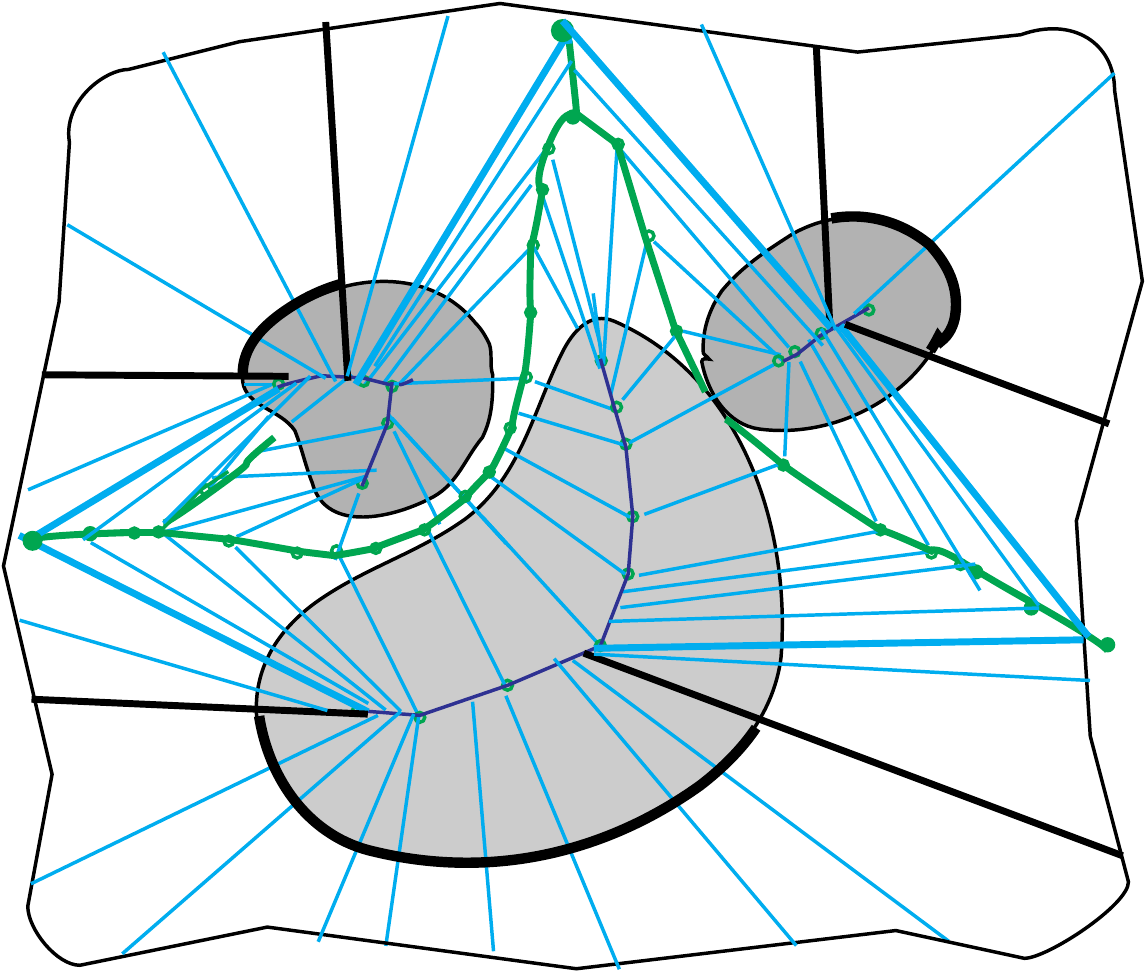}
\end{center}
(a) \hspace{2.5in} (b) \hspace {2.0in}  
\caption{\label{fig.II4.2} a) {\em Bounding box} with curved corners 
containing a configuration of three regions and b) (a priori given) {\em 
intrinsic bounding region} for the same configuration.  The linking 
structure is either extended to the boundary (in the region bounded by the 
darker lines on the boundary for $M_{i\, \infty}$) or truncated at the 
boundary.} 
\end{figure} 
\par
\vspace{1ex} \flushpar
{\it Possibilities for a Bounding Region $\tilde \gW$}:  \hfill \par
\vspace{1ex}
\flushpar
\begin{itemize}
\item[1)]{\it Bounding Box or Bounding Convex Region: } The box requires a 
center and directions and sizes for the edges of the box.  For this, we 
would need to first normalize the center and directions for the sides of 
the box and then normalize the sizes of the edges either using a fixed size 
or one based on feature sizes of the configuration (see e.g.  
Figure~\ref{fig.II4.2}~(a)).
\item[2)]{\it Convex Hull: } The smallest convex region which contains a 
configuration is the convex hull of the configuration.  For a generic 
configuration, the convex hull consists of the regions $\cB_{i\, \infty}$ 
together with the truncated envelope of the family of degenerate 
supporting hyperplanes which meet the configuration with a degenerate 
tangency or at multiple points. a) of Figure~\ref{fig.5b} in 
\S~\ref{S:sec3}, which is a configuration 
in $\R^2$, the envelope consists of line segments joining the doubly 
tangent points corresponding to the points in the spherical axis (b) of 
Figure~\ref{fig.5b}.  In $\R^3$, the envelope consists of a triangular 
portion of a triply tangent plane, with line segments joining pairs of 
points with a bitangent supporting plane, and a decreasing family of 
segments ending at a degenerate point.
\item[3)]{\it Intrinsic Bounding Region: } If the configuration is naturally 
contained in an (a priori given) intrinsic region, which is modeled by 
$\tilde \gW$, then provided that the extensions of the radial lines 
intersect $\partial \tilde \gW$ transversely, then we can modify the 
linking structure as in the convex case to have it defined on all of the 
$M_i$, and terminating at $\partial \tilde \gW$, if linking has not already 
occurred (see e.g. 
Figure~\ref{fig.II4.2}~(b)).
\item[4)]{\it Threshold for Linking: } The external region can be bounded by 
placing a threshold $\tau$ on the $\ell_i$ so that the $L_i$ remain in a 
bounded region.  This can be done in two different ways.  An {\it absolute 
threshold} restricts the external regions to those arising from linking 
vector fields with $\ell_i \leq \tau$; and a {\it truncated threshold} 
restricts to a bounded region formed by replacing $\ell_i$ by 
$\ell_i^{\prime} = \min\{\ell_i, \tau\}$.  For the 
first type, only part of the region would have an external linking 
neighborhood; while in the second, the entire region would.  As for the 
convex case, since we are replacing $\ell_i$ by smaller values, the linking 
flow will remain nonsingular.  We obtain modified versions of the regions 
lying in a bounded region. 
\end{itemize}

\section{Blum Medial Linking Structure for a Generic Multi-Region 
Configuration}
\label{S:sec3}
\par
In this section we consider two types of linking structures extending the 
Blum medial axes of the individual objects (i.e. regions).  To do so we give 
a number of results.  These include: giving the generic structure of the 
Blum medial axis for a region with singular boundary allowing corners and 
edges (in $\R^3$); introducing the \lq\lq spherical axis\rq\rq for the 
configuration, which plays the analogous role for directions as the medial 
axis plays for distances; and giving the generic properties for the \lq\lq 
full Blum medial linking structure\rq\rq for a general configuration, and 
its special form in the case of a configuration of disjoint regions with 
smooth boundaries.  \par

\subsection*{Blum Medial Axis for a Single Region with Generic Singular
Boundary} 
\par
\begin{figure}[ht] 
\centerline{\includegraphics[width=10cm]{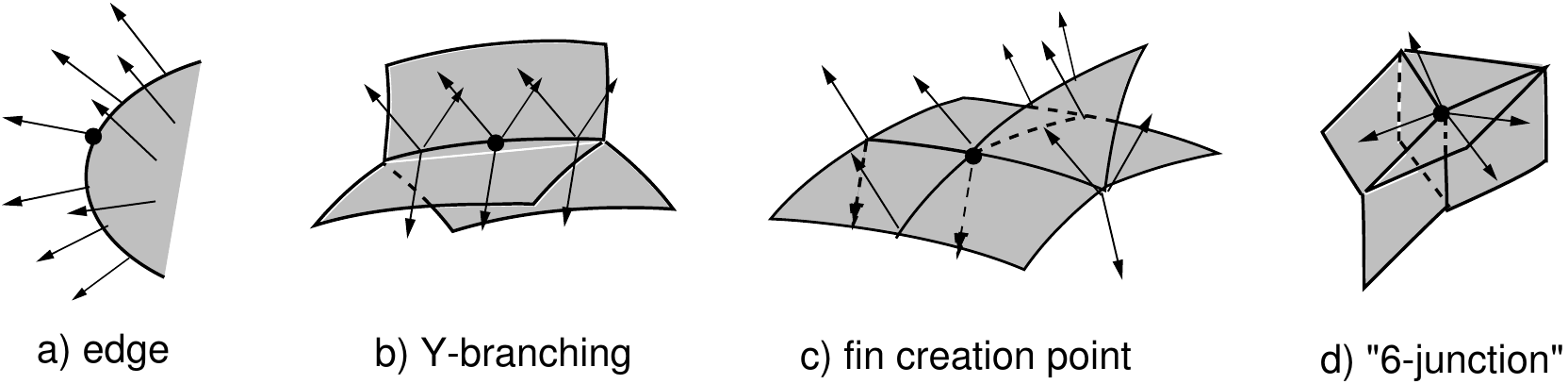}} 
\caption{\label{fig.3a} Generic local medial axis structures in $\R^3$: a) 
$A_3$; b) $A_1^3$; c) $A_1A_3$; and d) $A_1^4$.} 
\end{figure}

For a single region we will extend the classification of the generic local 
structure of the Blum medial axis for regions with smooth boundaries in 
$\R^2$ and $\R^3$ due to Yomdin \cite{Y}, Mather \cite{M}, and 
Giblin-Kimia \cite{GK} (see also \cite{P} and \cite{PS}).  For $\R^2$ the 
singular points are either 
$Y$-branch points where $3$ smooth curves meet non-tangentially at a 
point, or end points of a smooth curve.  For $\R^3$ the singular points are 
either edge points ($A_3$), fin points ($A_1A_3$), $Y$-branch curve 
points ($A_1^3$), or $6$-junction points ($A_1^4$), where six surface 
segments meet at a point along four $Y$-branch curves.  These are 
illustrated in Figure \ref{fig.3a}.  The notation $A_1A_3$ or $A_1^k$ 
denotes the singular behavior of the \lq\lq distance to the boundary\rq\rq 
function at the points on the boundary associated to the indicated points 
on the medial axis. 
\par
\par

\subsubsection*{Stratification of the Boundary} \hfill
\par
\begin{figure}[ht] 
\centerline{\includegraphics[width=10cm]{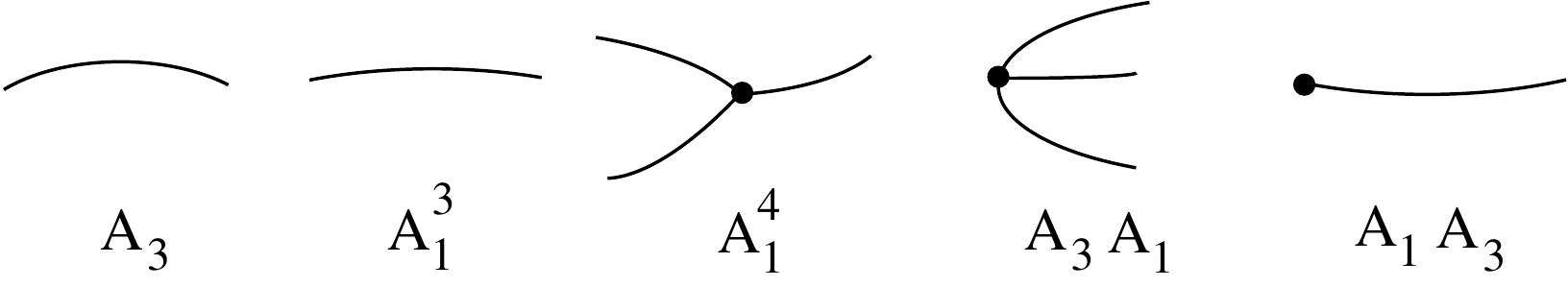}} 
\caption{\label{fig.3b} Generic local stratification of the boundary for a 
region in $\R^3$ in terms of the type on the medial axis corresponding to 
Figure~\ref{fig.3a}: $A_3$ and $A_1^3$ are the dimension $1$ strata 
(corresponding to a) and b)), while the other three are dimension 
$0$.  The last two represent $A_1A_3$ type from the locations of the 
$A_3$ point and the  $A_1$ point.} 
\end{figure}

The associated points on the boundary form a stratification of the 
boundary classified by the types of singular points.  The boundary 
classification is shown in \cite[Thm 4.4]{DG} to have the generic forms 
given in Figure \ref{fig.3b}.
\par
\par
\subsubsection*{Stratification of the Blum Medial Axis at Singular 
Boundary Points} \hfill
\par
\begin{figure}[ht]
\includegraphics[width=8cm]{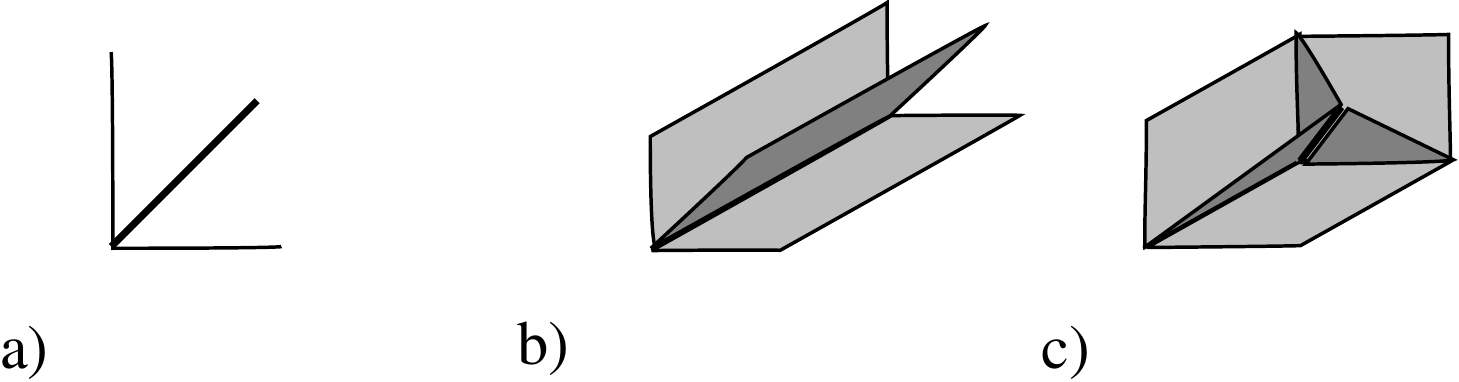}
\caption{\label{fig.3.3b} 
In $\R^2$, a) normal form for the Blum medial axis at a corner point on the 
boundary.  In $\R^3$, b), resp. c) are normal forms for edge points, resp. 
corner points.  The darker shaded regions represent the Blum medial axis.}  
\end{figure}

Unlike the case of a region with smooth boundary, the Blum medial axis of 
a region with singular boundary may extend up to the singular portion.  
However, in the case of singular points which are either corners or edges 
(in $\R^3$) we can give a precise characterization of the Blum medial axis 
near these points.  
\begin{Definition}
\label{Def3.3}
The {\it edge-corner normal form} for the Blum medial axis of 
an edge or corner point $x$ in $\R^2$ or $\R^3$ is the image via a smooth 
diffeomorphism, sending the origin to $x$, of one of the models given in 
Figure \ref{fig.3.3b}.  In $\R^2$, the model for the medial axis is given for 
the corner model by the set of points $(x, y)$ with $x = y \geq 0$.  In 
$\R^3$, the models are given by either: the set of points $(x, y, z)$ with $x 
= y \geq 0$ for the edge model, or by the set of points $(x, y, z)$ with $x, 
y, z \geq 0$ and one of $x = y \geq 0$, $x = z \geq 0$, or $y = z \geq 0$.  
\end{Definition}
\par
\par
We give the following strengthened form for the structure of the Blum 
medial axis allowing singular points, see \cite[Thm 4.4]{DG}
\begin{Thm}[Generic Properties of the Blum Medial Axis]
\label{Thm3.1}
For a generic region $\gW$ in $\R^2$ or $\R^3$ with singular boundary 
allowing corners and edges (for $\R^3$), the Blum medial axis $M$ has the 
following properties:
\begin{itemize}
\item[i)]  it has the same local properties in the interior of the region 
$\intr(\gW)$ as given above for smooth regions (including the local 
singular structure and properties of the radial vector fields as illustrated 
for 3D in Figure \ref{fig.3a});
\item[ii)] the corresponding stratifications of the boundary lie in the 
smooth strata (i.e. they miss the edges and corners) and are the same as 
for the smooth case; and
\item[iii)] the closure of the Blum medial axis in the boundary consists of 
the singular points of the boundary; at these points, the Blum medial axis 
has the corresponding edge-corner normal form in Definition \ref{Def3.3}.   
\end{itemize}
\end{Thm}
\par

\begin{Remark}
\label{Rem3.2}
The Blum medial structure $M$ for a region with boundary and corners 
contains the singular points of the boundary in its closure, and at such 
points the radial vector field $U = 0$.  Hence, $(M, U)$ does not define a 
skeletal structure in the strict sense.  However, it can still be used, just 
as with skeletal structures, to compute the local, relative, and global 
geometry and topology of both the region and its boundary.  Hence, we can 
view it as a \lq\lq relaxed skeletal structure,\rq\rq where \lq\lq 
relaxed\rq\rq means that $M$ includes the singular boundary points and $U 
= 0$ on these points.  Alternatively, we can modify the Blum medial axis 
near the singular points of the boundary so it becomes a skeletal 
structure, see \S \ref{S:sec4}. 
\end{Remark} 
\par
 
\subsection*{Spherical Axis of a Configuration} 
\par
The generic local properties of the Blum medial axis for individual regions 
described in the last section will also apply to the external linking axis 
for the Blum linking structure which will be intrinsic for a configuration.  
There are still two remaining contributions to the linking structure.  The 
one involves identifying the types of linking which may occur generically 
between different regions.  This leads to part of the refinement of the 
stratification of the Blum medial axes of the individual regions and the 
external linking axis.  The remaining contribution concerns the portions of 
the Blum medial axes, denote $M_{\infty}$, and their corresponding 
boundary points $\cB_{\infty}$ where no linking occurs.  These are points 
on the boundaries where no other regions  (or different portions of the 
same region) would be visible.  We characterize the  boundary of 
$\cB_{\infty}$ using the \lq\lq spherical axis\rq\rq and the associated 
\lq\lq spherical structure\rq\rq, which we now proceed to define.  
\par  
Along with the Blum medial axis, we introduce its analog for directions in 
place of distances.  Directions are given by vectors $\bv \in S^n$, where 
$S^n$ is the unit sphere in $\R^{n+1}$.  Given such a direction and either a 
region $\gW$ or a configuration $\bgW$, the {\it supporting hyperplane in 
the direction $\bv$} is given by an equation $x\cdot \bv = c$, where all 
points $x$ of $\gW$ or $\bgW$ satisfy $x\cdot v \leq c$ (i.e. it is 
contained in the half-space defined by $x\cdot \bv \leq c$), with equality 
at one or more points.  The intersection of the supporting hyperplanes 
defines the convex hull of the region or configuration.  \par
\begin{figure}[ht]
\begin{center}
\includegraphics[width=10cm]{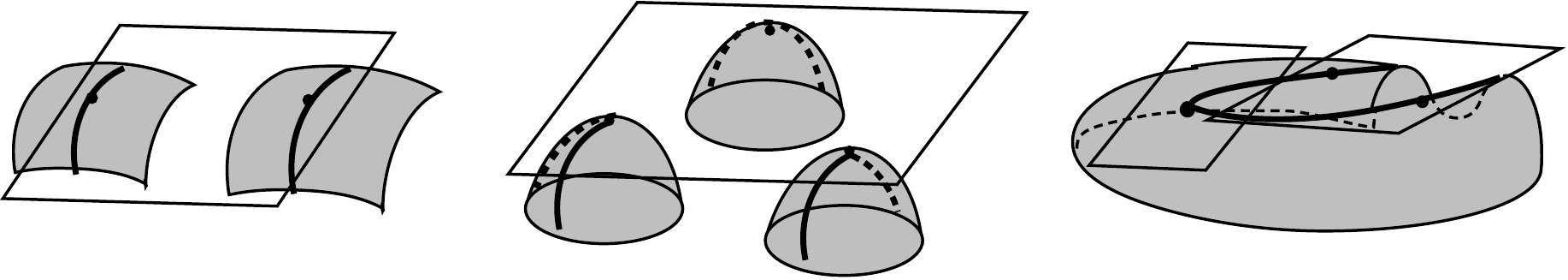} 
\end{center}
\hspace{0.5in} a) \hspace{1.0in} b) \hspace{1.5in} c) \hspace{0.5in}
\caption{\label{fig.3.3d}  Types of tangencies corresponding to points in 
the spherical axis.  The tangent planes shown are to points of multiple 
tangencies (except for the single $A_3$ point at the left tangency in c).}
\end{figure}

 
We define the {\it spherical axis} $\cZ \subset S^n$ of $\gW$, or the 
configuration $\bgW$, to consist of directions $\bv \in S^n$ for which the 
supporting hyperplanes $x\cdot \bv = c$ for the convex hull of $\gW$ or 
$\bgW$ have two or more tangencies with $\cB$ or a degenerate tangency, 
see e.g. Figure~\ref{fig.3.3d}.  
If we let $\tau : \cB \times S^n \to \R$ be defined by $\tau(x, \bv) = 
x\cdot \bv$, then $\tau$ is the family of
\lq\lq height functions\rq\rq\ on $\cB$.  The spherical axis is the set of 
$v \in S^n$ at which the absolute maximum of $\tau(\cdot, v)$ occurs at 
multiple points or is a degenerate maximum (this is the \lq\lq Maxwell 
set\rq\rq of $-\tau$ ).  Then, similar methods to those used by Mather can 
be applied to determine the generic structure of $\cZ$ (see \cite[Thm 
4.6]{DG}).  In the special cases of $\R^2$ and $\R^3$ they give the 
following.

\begin{figure}[hb] 
\begin{center}
\includegraphics[width=6cm]{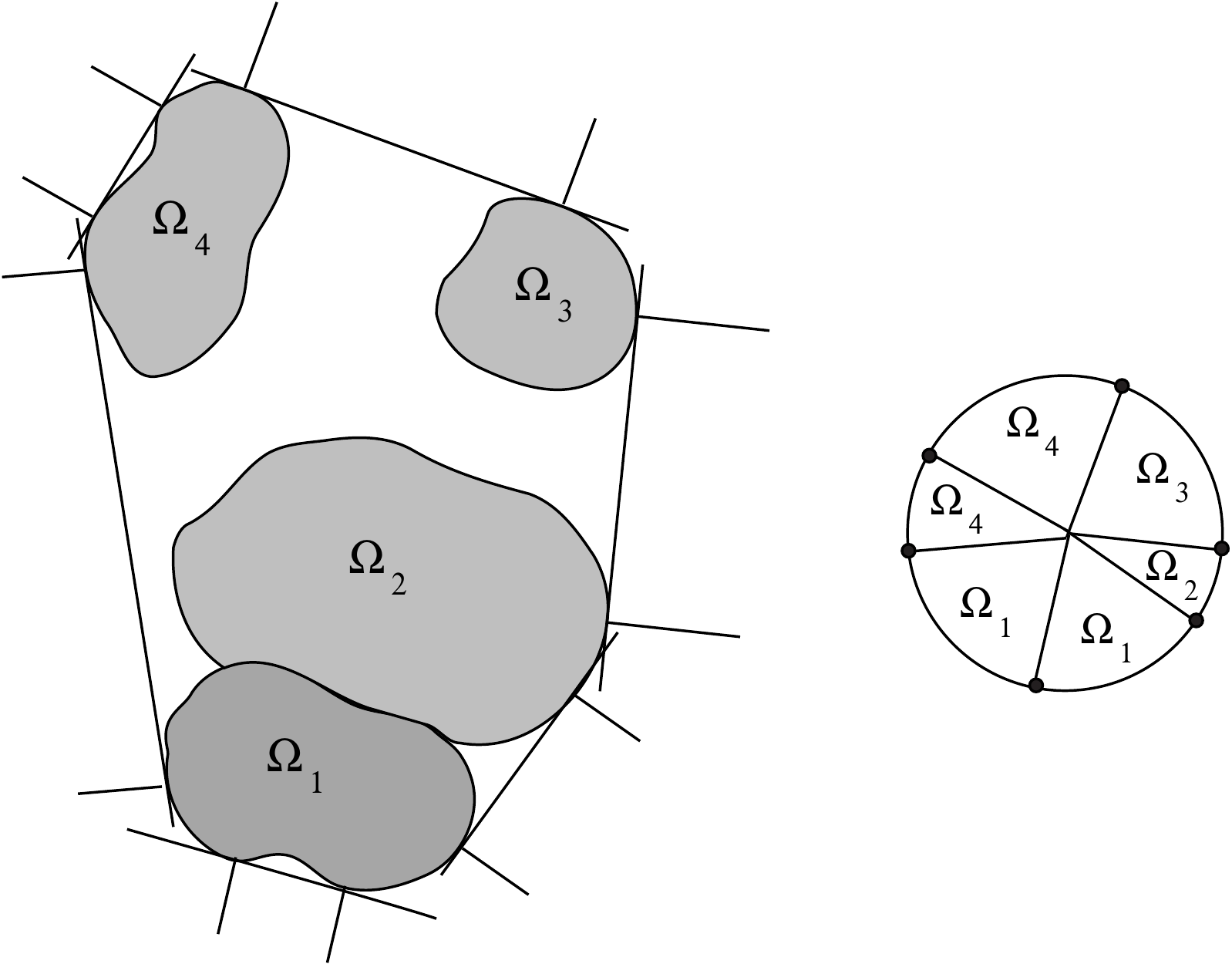} \hspace{1.5cm} 
\includegraphics[width=4cm]{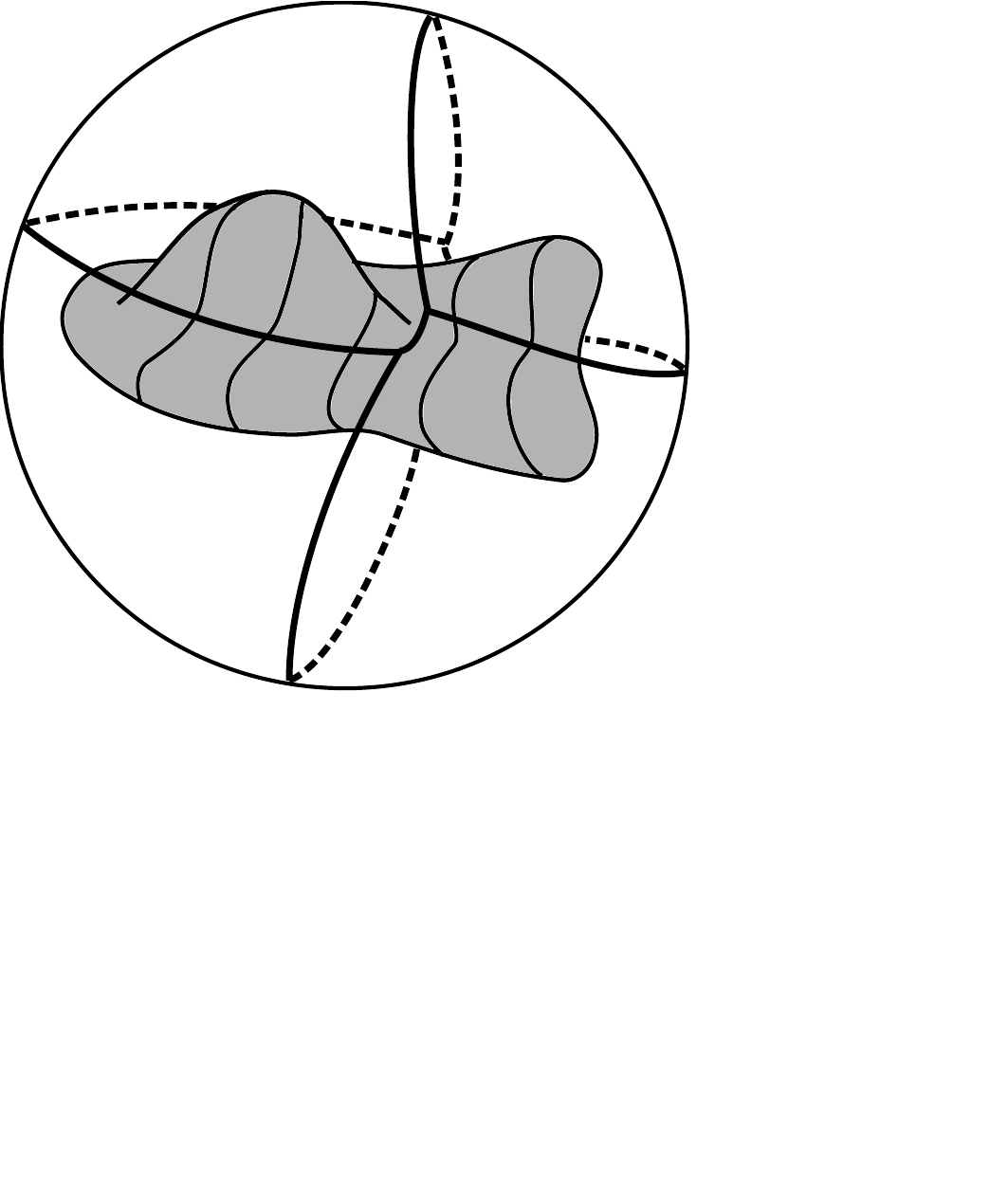} \\
 a) \hspace{1.5in} b) \hspace{1.5in} c)
\end{center} \flushpar
 \par
\caption{\label{fig.5b} a) Configuration of four regions in $\R^2$ with the 
bitangent supporting lines, and b) the corresponding spherical axis, which 
consists of the points on $S^1$ corresponding to bitangent lines in a).  The 
regions between points represent subregions of $\cB$ of unlinked points 
in $\cB_{\infty}$.  If the same region $\gW_i$ is indicated on both sides 
of a radial line, then in $\cB_i$ is a region involving self-linking.  The 
third figure in c) is a region in $\R^3$ with the corresponding spherical 
axis represented as a branched curve in the surrounding $2$-sphere.} 
\end{figure}

\begin{Thm}[Generic Properties of the Spherical Axis]
\label{Thm3.1b}
For generic regions (with smooth boundaries) or generic configurations, 
the spherical axis has the following local structure.  In $\R^2$, $\cZ$ 
consists of a discrete set of points where supporting lines have two 
tangent points (see e.g. Figure \ref{fig.5b}).  In $\R^3$, it consists of a 
collection of branched curves with endpoints.  At smooth points of the 
curves there is a double tangency of the supporting plane.  The branching 
occurs at $Y$-branch points which correspond to triple tangencies.  The 
end points correspond to points where the height function $\tau$ has an 
$A_3$ singularity.  These are illustrated in Figure \ref{fig.3.3d}.
\end{Thm}
\par
\par
\par 
\par

\subsubsection*{Spherical Structure for a Configuration}
\par
Just as for the Blum medial axis, we may associate to the spherical axis 
$\cZ$, both a height function $h$ and a multivalued vector field $V$.  To do 
this we need to initialize the origin.  Then, the height for a point $\bu$ on 
the spherical axis, which is a unit vector, defined for a point $x \in \cB$ 
is just the dot product $x \cdot \bu$.  Furthermore, there may be multiple 
points associated to $\bu$, all of which lie in the supporting hyperplane 
$H$ defined by $x \cdot \bu = h(\bu)$, for the maximum value $h = h(\bu)$ 
for $\cB$.  For each point $x_i \in H \cap \cB$, there is a vector 
$V = x_i - h\bu$ orthogonal to the line spanned by $\bu$.  This defines a 
multivalued vector field $V$.  \par
\begin{Definition}
\label{Def3.2b} The full {\em spherical structure} for the spherical axis of 
the multi-region configuration is the triple $(\cZ, h, V)$, consisting of the 
spherical axis $\cZ$, the height function $h$, and the multivalued vector 
field $V$.  This depends upon the choice of an origin on which all height 
functions are $0$. 
\end{Definition}
From the spherical structure, we can reconstruct the boundary of 
$\cB_{\infty}$ by $x = V(\bu) + h(\bu) \bu$ for $\bu \in \cZ$ and the 
multiple values of $V$ at $\bu$.  Here $x$ denotes a collection of points 
corresponding to the values of $V$.  \par 
Then, the regions in $\intr(\cB_{\infty})$ are the regions in the 
complement of the boundary of $\cB_{\infty}$ which have supporting 
hyperplanes for at least one point in one of the corresponding 
complementary regions to the spherical axis.  If we have in addition the 
height function for the configuration defined on all of $S^n$, then we can 
construct the supporting hyperplanes for all $\bu \in S^n$, and the 
envelope of these hyperplanes yields $\cB_{\infty}$.  
\par
\subsection*{Blum Medial Linking Structure} 
\par
We now consider the analogous Blum medial linking structure for a generic 
multi-region configuration.  First, we note that if the configuration has 
regions with boundaries and corners, then the Blum medial axes of the 
individual regions will not define skeletal structures. This is because the 
Blum medial axis will actually meet the boundary at the edges and 
corners.  However, in the case of disjoint regions $\{\gW_i\}$ in 
$\R^{n+1}$ with smooth generic boundaries (which do not intersect on 
their boundaries) there is a natural Blum version of a linking structure, 
which we introduce.  \par

\begin{Definition} Given a multi-region configuration of disjoint regions 
$\bgW = \{\gW_i\}$ in $\R^{n+1}$, for $i = 1, \dots , m$, with smooth 
generic boundaries (which do not intersect on their boundaries), a {\em 
Blum medial linking structure} is a skeletal linking structure for which: 
\begin{itemize} 
\item[B1)] the $M_i$ are the Blum medial axes of the regions $\gW_i$ 
with $U_i$ the corresponding radial vector fields;
\item[B2)] the linking axis $M_0$ is the Blum medial axis of the exterior 
region $\gW_0$ (and we refer to it as the {\em medial linking axis}); 
and
\item[B3)] the $M_{i\, \infty}$ are the points in $\tilde M_i$ 
corresponding to points on $\cB_i$ for which a height function has an 
absolute maximum (or minimum for the height function for the opposite 
direction).  
\end{itemize} 
\end{Definition}
\par
\begin{Remark}
It follows from B2) that if $x \in M_i$ and $x^{\prime} \in M_j$ are linked, 
the corresponding values of the radial and linking functions satisfy 
$\ell_i(x) - r_i(x) = \ell_j(x^{\prime}) - r_j(x^{\prime})$. 
\end{Remark}
\vspace{1ex}
\subsection*{\it Generic Linking Properties} \hfill
\par
We consider a generic configuration of regions $\{ \gW_i\}$ with $\cB_i$ 
the boundary of $\gW_i$, and $M_i$ the Blum medial axis of $\gW_i$.  If 
each region is generic, then by Theorem \ref{Thm3.1} the Blum medial axis 
of each region has generic local structure.  For 3D, the points on the 
boundary are of types $A_3$, $A_1^3$, $A_1^4$, $A_3A_1$ and $A_1A_3$, 
as shown in Figure \ref{fig.3b}, along with the remaining points of type 
$A_1^2$ corresponding to smooth points on the medial axis.  For 2D, there 
are only isolated points of type $A_3$, corresponding to end points of the 
medial axis and $A_1^3$ points  corresponding to $Y$-branch points, with 
the remaining points of type $A_1^2$.  
\par
\begin{figure}[h] 
\centerline{\includegraphics[width=7cm]{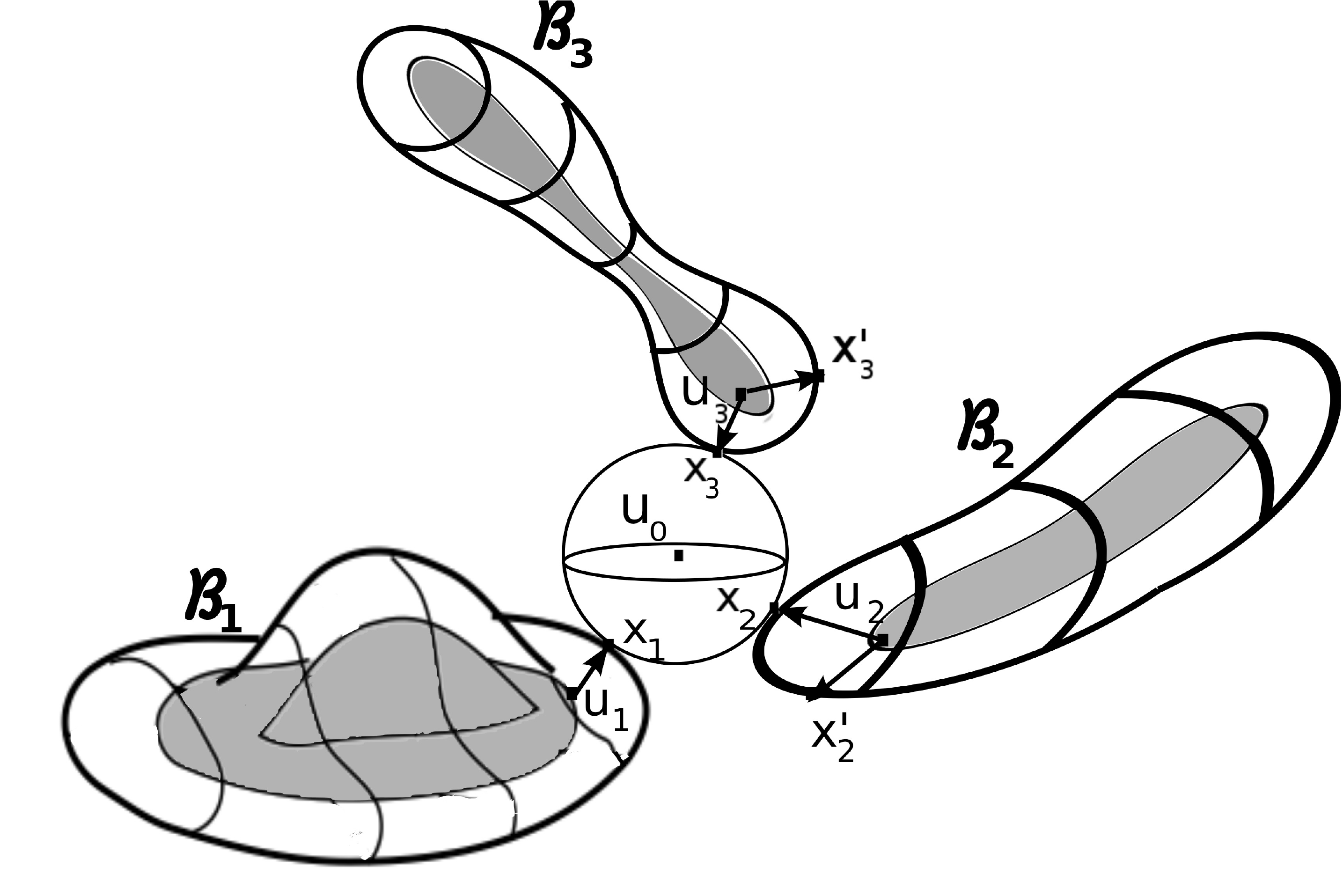}} 
\caption{\label{fig.5e} Generic linking involves the simultaneous role 
played by points $x_i \in \cB_i$ and their Blum features via the distance 
function to a point $u_i$ of the Blum medial axis $M_i$ of the interior 
region $\gW_i$ and that for a point $u_0$ in the external linking medial 
axis .} 
\end{figure}
\par

A point of $\cB_i$ may simultaneously have a role as both a point on the 
boundary of $\gW_i$ and as a point on the boundary of the external region, 
see Figure \ref{fig.5e}.  As such it has two distinct Blum descriptions, and 
hence simultaneously belongs to strata on the boundaries corresponding to 
given Blum behavior exhibited for the individual regions and for the 
external region.  Generic linking between regions for a configuration 
satisfies special genericity properties which is explained by how these 
strata intersect.  
\par
We consider a collection of smooth boundary points $S = \{x_1, \dots, 
x_k\}$ with $x_i \in \cB_{j_i}$ associated to points $u_i$ on the medial 
axes $M_{j_i}$.  Suppose the $u_i$ are linked at the point $u_0$ on the 
external medial axis $M_0$.  In the generic case, this means $k \leq 3$ for 
$\R^2$ and $k \leq 4$ for $\R^3$.  We use the notation $A_{\bga_i}$ to 
denote the Blum type of $x_i$ for the point $u_i$ for the individual region 
$\gW_{j_i}$ and denote the corresponding strata of the boundary 
consisting of points of this type by $\gS^{\bga_i}_{\cB_{j_i}}$.  Likewise, 
we let $A_{\bgb}$ denote the Blum type for the point $u_0$ on the 
external medial axis, and denote the strata in the external medial axis by 
$\gS_{M_0}^{(\bgb)}$ and the corresponding strata on the boundary by  
$\gS^{\bgb}_{\cB}$.  Then, the {\it Blum linking type} is denoted by 
$(A_{\bgb}: A_{\bga_{i_1}}, \dots , A_{\bga_{i_k}})$. 
\par
\begin{figure}[ht]
\begin{center}
\includegraphics[width=9.0cm]{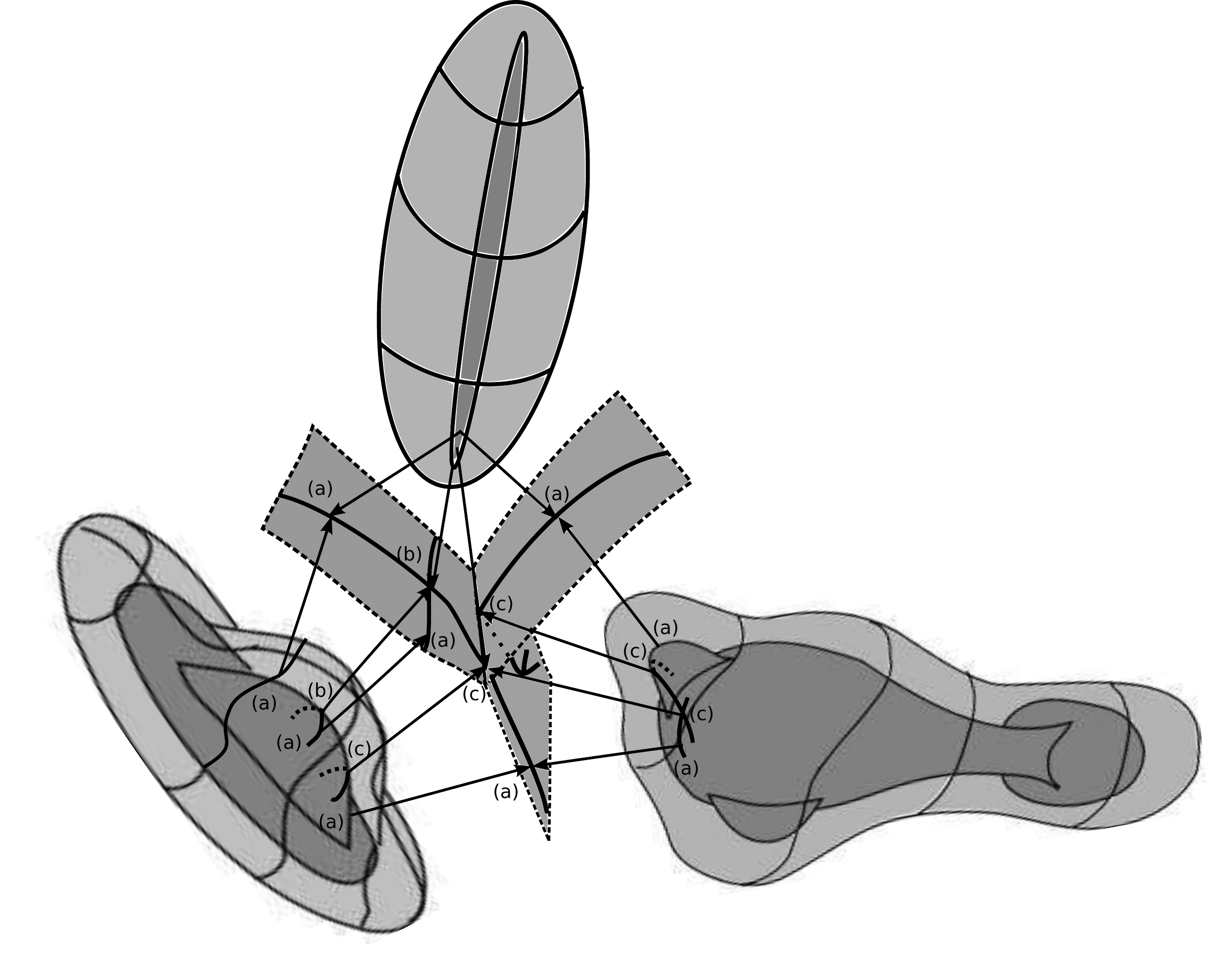}
\end{center} 
\caption{\label{fig.3.3c} Examples of linking types involving three objects 
with a portion of the linking medial axis shown.  The points (a) are of 
linking type $(A_1^2: A_3, A_1^2)$; points (b) illustrate linking of type 
$(A_1^2: A_3, A_3)$, and (c) of type $(A_1^3: A_3, A_1^2, A_1^2)$.  In the 
external linking medial axis is illustrated via the dark curves the 
$1$-dimensional strata of type $(A_1^2: A_3, A_1^2)$ with isolated points 
either of types $(A_1^2: A_3, A_3)$ where the $1$-dimensional strata 
cross or $(A_1^3: A_3, A_1^2)$ where the $(A_1^2: A_3, A_1^2)$ curve 
crosses the $Y$-branch curve of the external linking medial axis.}
\end{figure}

\begin{Definition}
\label{Def3.4}
The set of points $S = \{x_1, \dots, x_k\}$ as above exhibits {\em generic 
Blum linking} of type $(A_{\bgb}: A_{\bga_{i_1}}, \dots , A_{\bga_{i_k}})$ 
if: 
\begin{itemize}
\item[i)]  the strata $\gS_{\cB}^{(\bgb)} \subset \cB$ and 
$\gS_{\cB_{j_i}}^{(\bga_i)} \subset \cB_{j_i}$ intersect transversely in 
$\cB_{j_i}$; and 
\item[ii)] the images in $M_0$ of the strata $\gS_{\cB}^{(\bgb)} \cap 
\gS_{\cB_{j_i}}^{(\bga_i)}$ under the linking flow intersect transversely 
in the stratum $\gS_{M_0}^{(\bgb)}$ (see e.g. Figure~\ref{fig.3.3c}). 
\end{itemize}
\end{Definition}
\par
For example, generic linking in $\R^2$ means that zero-dimensional 
(isolated point) strata in one region can only be linked to points in the 
$A_1^2$ strata in another.  Also, in $\R^3$ a one-dimensional stratum can 
intersect other one-dimensional stratum (transversely in a single point) 
in a boundary $\cB_i$ and this can then also be linked to $A_1^2$ strata in 
another boundary as in Figure \ref{fig.3.3c}.  \par 
\begin{Example}[Generic Linking in $\R^2$]
\label{2Dlink}
There are only five types of generic linking for configurations in $\R^2$ 
(see Figure~\ref{fig.5d}):  
\begin{itemize} 
\item[i)] $(A_1^2 : A_1^2, A_1^2)$:\, linking smooth points of two medial 
axes at a smooth point of the external medial axis, forming 
one-dimensional strata; 
\item[ii)] $(A_1^2 : A_3, A_1^2)$:\, linking an end point of one medial axis 
with a smooth point of another at an isolated smooth point of the external 
medial axis, forming a zero-dimensional stratum;
\item[iii)] $(A_1^2 : A_1^3, A_1^2)$:\, linking a $Y$-branch point of one 
medial axis with a smooth point of another at an isolated smooth point of 
the external medial axis, forming a zero-dimensional stratum;
\item[iv)] $(A_1^3 : A_1^2, A_1^2, A_1^2)$:\, linking smooth points of 
three medial axes at a $Y$ branch point of the external medial axis, 
forming a zero-dimensional stratum;
\item[v)] $(A_3 : A_1^2)$:\, a single smooth point of a medial axis is 
self-linked at an end point of the external medial axis.
\end{itemize}
We note that all five types can occur for self-linking; but only the first 
four can occur for linking between distinct regions; and partial linking can 
only occur via iv).  \par 
By these being the only five generically occurring linking types, we mean 
that any other possibility, such as e.g. two end points of different regions 
being linked, may be arranged for a specially constructed configuration; 
however, almost all slight perturbations of this special configuration will 
destroy this special linking feature and only involve the five basic types. 
\end{Example}
\par
For 3D, the situation becomes more complicated and there are seventeen 
generic linking types listed in Table \ref{Table1} in \S \ref{S:sec6}.  
However, as explained in \S \ref{S:sec7}, by replacing the Blum linking 
structure with a simpler skeletal linking structure, we can significantly 
reduce the number of linking types. 

\par
\par

\begin{Remark}
\label{Rem3.5}
For a general multi-region configuration, we can substitute in place of 
$\gW_0$ a region $\gW_i$ which has multiple adjoining regions (including 
possibly the complement $\gW_0$) and the definition of \lq\lq generic 
linking\rq\rq of the adjoining regions relative to $\gW_i$ has the same 
form as in Definition \ref{Def3.4}.  Generically they have 
the same properties as for $\gW_0$. \par
\end{Remark}

\subsection*{\it Generic Structure for $\cB_{\infty}$ and $M_{\infty}$}  
\hfill
\par
For a region $\gW_i$ of a configuration, the points in $M_{i\, \infty}$ (and 
$\cB_{i\, \infty}$) are not involved in linking; however, they do involve 
points which have a type $A_{\bga}$ based on the internal medial 
structure of $\gW_i$.  By the generic structure of $M_{i\, \infty}$ and 
$\cB_{i\, \infty}$ we are interested in both the generic properties of the 
stratifications of $\cB_{i\, \infty}$ and $M_{i\, \infty}$ resulting from 
the spherical axis and their relation with the strata $\gS^{\bga}_{\cB_i}$.   
\par
The interior points of $\cB_{i\, \infty}$ are those points where the 
supporting hyperplane meets the configuration at a single nondegenerate 
tangency.  In addition, the boundary of $\cB_{i\, \infty}$ 
consists of strata $\gS^{(\bga)}_{\infty}$ corresponding to the strata 
$\gS^{(\bga)}_{\cZ}$ of the spherical axis $\cZ$ and are of the same 
dimensions.  The strata $\gS^{(\bga)}_{\infty}$ lie in the smooth strata of 
the $\cB_i$.  

\begin{Definition}
\label{Def3.4b}
By $M_{\infty}$ and $\cB_{\infty}$ having {\em generic structure} we 
mean:
\begin{itemize}
\item[1)] the spherical axis of the configuration is generic and each 
$\cB_{i\, \infty}$ has strata with the resulting generic local structure 
given by Theorem \ref{Thm3.1b}; 
\item[2)] the strata of this stratification of $\cB_{i\, \infty}$ intersect 
tranversally the strata $\gS^{(\bgb)}_{\cB_i}$ for the Blum medial 
axis of $\gW_i$; and 
\item[3)]  the strata of $M_{i\, \infty}$ are the images in $\tilde M_i$ of 
the transverse intersections 
$\gS^{(\bga)}_{\infty} \cap \gS^{(\bgb)}_{\cB_i}$. 
\end{itemize}
\end{Definition}
\par
Examples of the generic local structure for the boundary strata of 
$M_{\infty}$ for a multi-region configuration in $\R^3$ are shown in 
Figure \ref{fig.3.3d}: a) of type  $A_1^2$, b) of type $A_1^3$, and c) of type 
$A_3$.  The darker curves (including the darker dashed curves) denote the 
boundary strata bounding regions of $M_{\infty}$ (consisting of points 
whose outward pointing normals point away from the other regions).
\subsection*{Existence of a Blum Medial Linking Structure} 
\par
Before stating the general form of a full Blum linking structure for a 
general configuration, we first give a special case where the 
configuration $\bgW$ consists of disjoint regions with smooth boundaries.  
Then, the existence of Blum medial linking structures is guaranteed by the 
following, which in addition ensures generic linking (see \cite[Thms 3.4.3 
and 3.4.4]{Ga} and \cite[Thm 4.12]{DG}).

\begin{Thm}[Existence of Blum Medial Linking Structure]
\label{Thm3.5}
For a generic configuration $\bgW = \{\gW_i\}$ in $\R^2$ or $\R^3$ 
consisting of disjoint regions with smooth boundaries (i.e. they do not 
intersect on their boundaries), and with $\bgW$ contained in the interior 
of a given compact region $\tilde \gW$,
\begin{enumerate}
\item  the configuration has a Blum medial linking structure such that 
each $M_i$ (including $M_0$) has generic local properties given by 
Theorem \ref{Thm3.1};
\item  the linking structure exhibits generic linking as in Definition 
\ref{Def3.4}; 
\item  $M_{\infty}$ and $\cB_{\infty}$ have generic structure as given in 
Definition \ref{Def3.4b}; and 
\item in the case that $\tilde \gW$ is convex, the properties for a linking 
structure in the bounded case hold.  
\end{enumerate}
\end{Thm}
\par
	This result is a special case of the following general result for a 
general configuration allowing adjoined regions in \cite[Thm 4.13]{DG}. 
\begin{Thm}[Full Blum Linking Structures]
\label{Thm3.6}
A generic multi-region configuration $\bgW = \{\gW_i\}$ in $\R^2$ or 
$\R^3$ contained in the interior of a compact region $\tilde \gW$ has the 
following generic properties: 
\begin{itemize}
\item[i)] each $\gW_i$ has a Blum medial axis $M_i$ exhibiting the {\em 
generic local properties at interior points} of $\gW_i$ given by Theorem 
\ref{Thm3.5}; 
\item[ii)] the {\em external region} in $\tilde \gW$ has a medial linking 
axis $M_0$ which {\em exhibits the generic local Blum properties};  
\item[iii)] the local structure of the Blum axes $M_i$ (including $i = 0$) 
near a singular boundary point has the {\em local generic edge-corner 
normal form} given by Definition \ref{Def3.3};
\item[iv)]  at a smooth boundary point of a region $\gW_i$ of type $Q_k$, 
$k = 2, 3$, the strata of $Q_k$-points transversally intersects the strata 
of Blum type points $\gS_{\cB_i}^{(\bga)}$;
\item[v)] {\em generic linking occurs between the smooth points} of the 
regions and no linking occurs at edge-corner points; and
\item[vi)]  $\cB_{i\, \infty}$ is contained in the smooth strata of the 
$\cB_i$, and {\em $\cB_{i\, \infty}$ and $M_{i\, \infty}$ exhibit the 
generic properties} given in Definition \ref{Def3.4b}.
\end{itemize}  
\end{Thm}
Note:  Property v) holds as well for generic linking between adjoining 
regions of a given region $\gW_i$ relative to the region $\gW_i$.  
\par

\section{Modifying the Full Blum Medial Structure to a Skeletal Linking 
Structure}
\label{S:sec4}
We know by Theorem \ref{Thm3.6} that a generic multi-region 
configuration has a Blum medial structure.  If the regions are disjoint 
with smooth boundaries, then the Blum linking structure is a skeletal 
linking structure.  However, if the configuration contains regions which 
adjoin, then the Blum linking structure does not satisfy all of the 
conditions for being a skeletal linking structure.  Specifically the 
individual Blum medial axes of both the regions and the complement will 
extend to the singular points of the boundaries. 
\par
There are two perspectives on this.  On the one hand, as mentioned in 
Remark \ref{Rem3.2}, we may view this as a \lq\lq relaxed form of a 
skeletal linking structure.\rq\rq  We shall see that from this structure we 
still obtain all of the local, relative, and global geometry of the individual 
regions and the positional geometry of the configuration.  However, if we 
consider the stability and deformation properties, such a structure does 
not behave well.  \par 
We describe two approaches to modifying the full Blum linking structure 
to obtain a skeletal linking structure.  One approach is when the 
configuration with adjoined regions can be viewed as a deformation of a 
configuration with disjoint regions.  The second is to modify the full Blum 
linking structure by a process of \lq\lq smoothing the corners\rq\rq of 
the regions.  We describe each of these.  \par

\subsection*{Example of Evolving Skeletal Linking Structure for Simple 
Generic Transition} \par
\begin{figure}[ht]
\begin{center} 
\includegraphics[width=8cm]{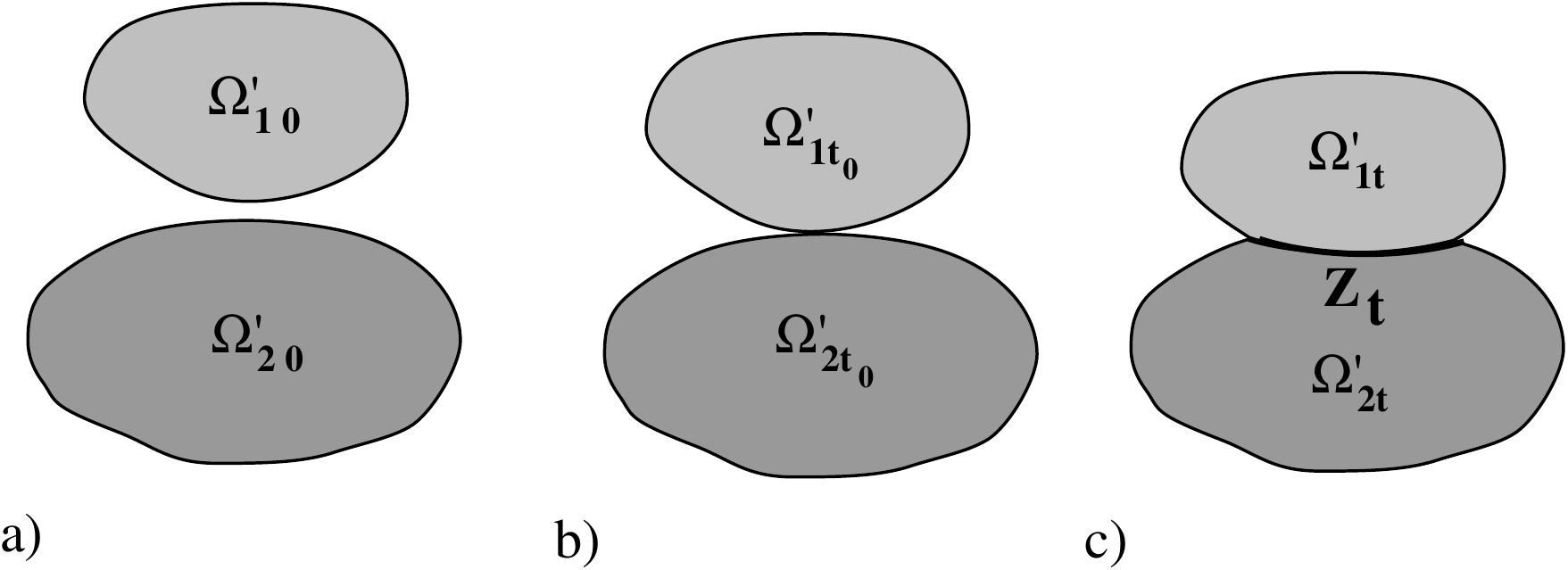} 
\end{center} 
\caption{\label{fig.8.1} The stages for a simple generic transition of two 
evolving regions $\gW^{\prime}_{i\, t}$ becoming adjoined: a) disjoint 
regions, b) simple tangency at $t = t_0$, and c) regions adjoined along 
$Z_t$ at $t > t_0$.}  
\end{figure}

We illustrate the method for a {\it simple generic transition} under which 
two initially disjoint regions $\gW_{i\, 0}$, $i = 1, 2$, undergo a 
deformation $\gW_{i\, t}$ to become adjoined at $t = t_0$ and form a 
common boundary region $Z_t$ for $t > t_0$.  
This is illustrated in Figure~\ref{fig.8.1}.  
\par
\par
\begin{figure}[ht]
\begin{center} 
\includegraphics[width=8cm]{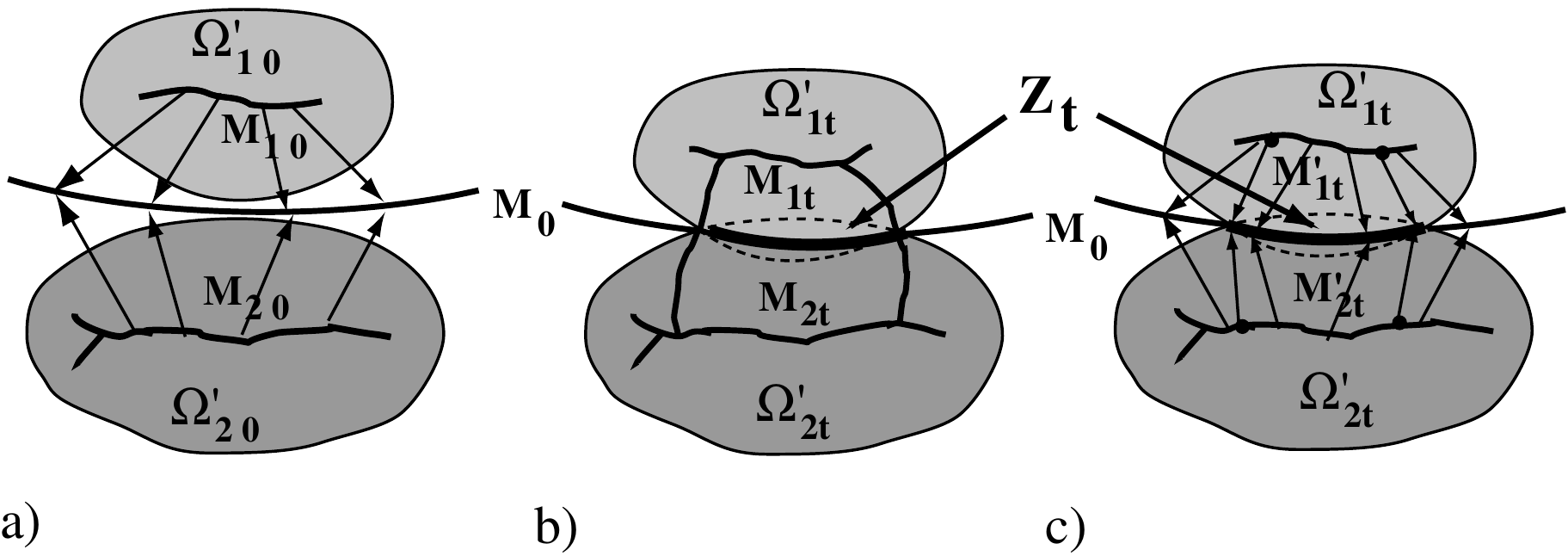} 
\end{center} 
\caption{\label{fig.8.2} Comparison of generic bifurcation of the full Blum 
linking structure versus evolution of the retracted skeletal linking 
structure for two evolving regions $\gW^{\prime}_{i\, t}$ becoming 
generically adjoined as in Figure~\ref{fig.8.1}.  Full Blum linking structure 
bifurcates by adding branches from a) unjoined regions to b) after 
becoming adjoined.  By contrast, the retracted skeletal linking structure 
evolves while retaining the structure of the skeletal sets, from a) 
unjoined regions to c) after being adjoined.}  
\end{figure}

However, this causes a transition in the full Blum medial linking structure 
as in changing from a) to b) in Figure~\ref{fig.8.2}.  There is a 
discontinuous change in the Blum medial axis somewhat analogous to the 
introduction of a $C^1$ bump in the boundary of a region which forces the 
Blum medial axis to introduce a new branch.  \par
An alternate approach is to deform the Blum medial linking structure of 
the regions before the transition to a skeletal linking structure as the 
transition occurs and continues for the adjoined regions to evolve from a) 
to c) in Figure~\ref{fig.8.2}.  This is achieved by keeping the skeletal axes 
for each region but altering the lengths of both the radial vectors so they 
extend to the common boundary and altering the linking vectors to the 
external linking axis as shown.  
\par 
This gives a family of skeletal linking structures for the varying 
configuration $\bgW_{t}^{\prime} = \{\gW_{1\, t}, \gW_{2\, t}\}$, which 
evolve continuously (and stratawise differentiably on the added strata 
$S_{i\, t} $), see Figure~\ref{fig.8.2}.
\begin{Remark}
\label{Rem5.1}
This example illustrates one significant advantage of skeletal linking 
structures over the full Blum linking structure for general multi-object 
configurations; namely, certain changes in configuration type can be 
incorporated as continuous variation in the skeletal linking structure. 
\end{Remark}
\par
\subsection*{Modifying the Full Blum Linking Structure via Smoothing} 
\par
\begin{figure}[ht]
\begin{center} 
\includegraphics[width=8cm]{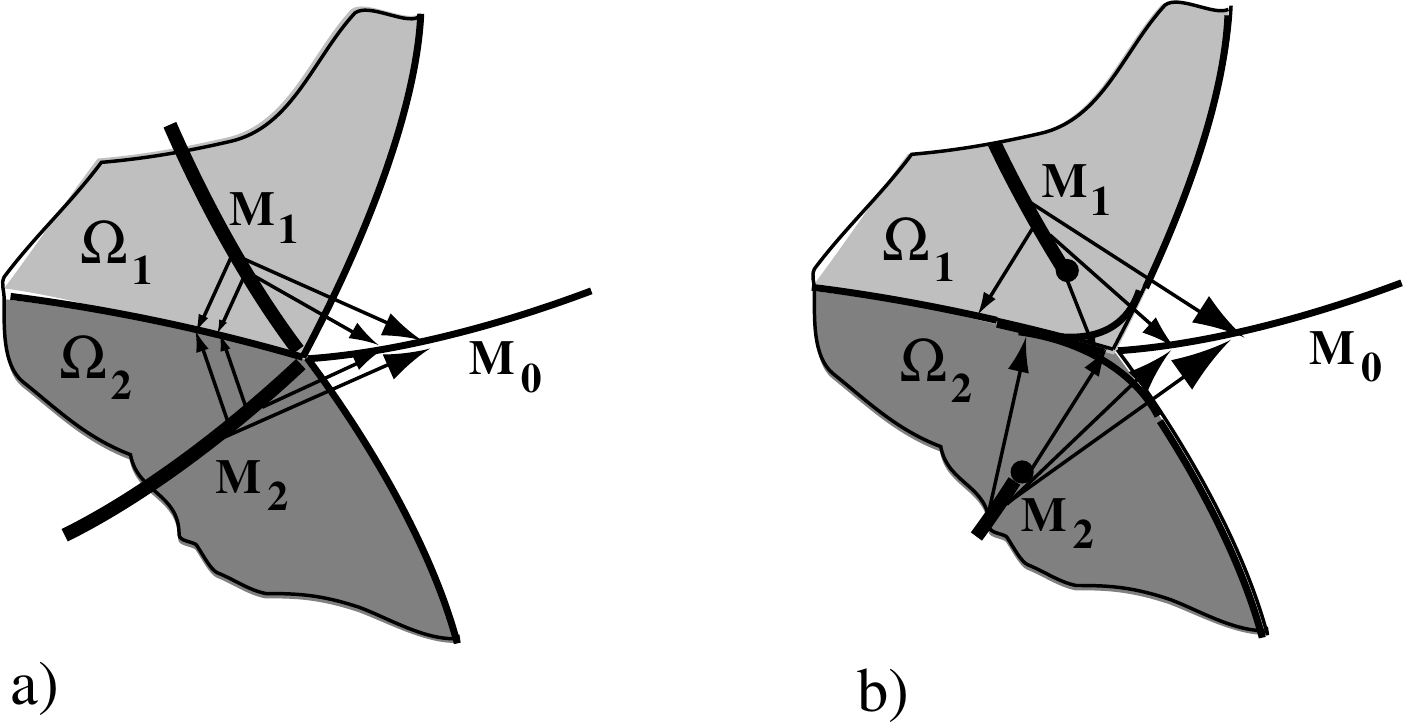} 
\end{center} 
\caption{\label{fig.8.3} An example of a smoothing of a configuration with 
adjoining regions in the neighborhood of a corner point:  
a)  Blum linking structure and b) Smoothing and resulting modified 
skeletal linking structure.}  
\end{figure}

A second approach for a configuration with adjoining regions is to modify 
the full Blum linking structure to a skeletal linking 
structure.  One way to accomplish this is by \lq\lq smoothing of the 
corners of the regions\rq\rq in a small neighborhood of the edge/corner 
set as in Figure \ref{fig.8.3}.  The goal is to modify the regions $\gW_i$ in 
a small neighborhood $W$ of the edge corner points 
 so that the smooth region boundaries are smooth, interior to $\gW_i$ and 
agree with the original boundaries outside of $W$.  This is done in such a 
way that the radial vector field of the Blum medial axes of the smooth 
regions can be extended to be transverse to the original boundary and can 
be extended to meet the external medial axis transversely.  These 
extended vectors are then the linking vectors for a skeletal structure.  The 
details of this can be found in \cite[Chap. 5]{DG}.
\par
\par

\section{Classification of Linking Types for Blum Medial 
Linking Structures in $\R^3$} 
\label{S:sec6}
\par
For a generic configuration $\bgW$ in $\R^3$, the Blum medial linking 
structure exhibits generic linking properties given in Table 
\ref{Table1}.  We first briefly explain the features of the table.  
Dimension refers to the dimension in $\R^3$ of the strata where the given 
linking type occurs.  There are three 
types of linking: 1) linking between points on distinct medial axes; 2) 
\lq\lq partial linking\rq\rq\, involving more than one point from one 
medial 
axis and point(s) from another; and 3) \lq\lq self-linking\rq\rq\, where 
linking is between points from a single medial axis.  \lq\lq Pure linking 
type\rq\rq\, refers to cases only occurring for self linking.  The $A_1^2$ 
and $A_1A_3$ linking can occur for either linking or self-linking; while 
$A_1^3$ and $A_1^4$ linking can occur for any of the three linking types.

\par
\begin{longtable}{lccl}
\caption{Classification of Linking Types for Blum Medial Linking 
Structures in $\R^3$} \\
  &  Linking Type  &   Dimension   &   Description of Linking  \\  
\hline
\endhead
   &  {\bf $A_1^2$ Linking}  &  &   \\ 
 [1ex] 
i) & $(A_1^2 : A_1^2, A_1^2)$   &   2   &   between $2$ smooth points \\  
[1ex]
ii) & $(A_1^2 : A_1^3,  A_1^2)$  &   1   &   between a smooth point \\
  &  &     & and a Y-junction point \\  [1ex]

iii) &  $(A_1^2 : A_3,  A_1^2)$  &  1  &   between a smooth point  \\
  &   &    &   and an edge point   \\  [1ex]

iv) &  $(A_1^2 : A_3A_1,  A_1^2)$   &  0   & between a fin point \\
  &   &     &   and a smooth point  \\   [1ex]

v) &  $(A_1^2 : A_1A_3,  A_1^2)$   &  0   & between a smooth point  \\
  &   &     &  associated to a fin point and \\
  &    &    &    another smooth point  \\   [1ex]

vi) &  $(A_1^2 : A_1^4,  A_1^2)$   &  0   &  between a smooth point \\
  &   &    &  and a $6$-junction point \\   [1ex]

vii) & $(A_1^2 : A_1^3,  A_1^3)$ &  0  & between $2$ Y-junction points \\
   &    &     &  \\  [1ex]

viii) & $(A_1^2 : A_3,  A_3)$ &  0  & between $2$ edge points \\
   &    &     &   \\  [1ex] 
ix) &  $(A_1^2 : A_1^3,  A_3)$ &  0  & between a Y-junction point  \\
   &   &     &   and an edge point \\  [2ex]

 $\text{ }$  & {\bf $A_1^3$, $A_1^4$ and $A_1A_3$ Linking}  &   &  \\  
[1ex]

x)  &  $(A_1^3 : A_1^2,  A_1^2, A_1^2)$   &  1   &  between $3$ smooth 
points \\   [1ex]
 xi)  &  $(A_1^3 : A_1^3,  A_1^2,  A_1^2)$  &   0   &  between $2$ smooth 
points \\
   &  &     & and a Y-junction point \\ [1ex]

xii)  &  $(A_1^3 : A_3,  A_1^2,  A_1^2)$  &  0  &  between $2$ smooth 
points  \\
   &  &    &   and an edge point   \\  [1ex]

xiii)  &  $(A_1^4 : A_1^2,  A_1^2, A_1^2, A_1^2)$   &  0   &  between $4$ 
smooth points \\   [1ex]

xiv) &  $(A_1A_3 : A_1^2, A_1^2)$   &  0  &  $A_1A_3 $ linking between 
$2$\\ [1ex]     &  &    &   smooth points  \\ [2ex]

   & {\bf Pure Self-Linking}  &  &  \\ [1ex]

xv)  &  $(A_3 : A_1^2)$   &  1   & edge-type self-linking with \\
   &   &     &   a smooth point \\  [1ex]

xvi)  &  $(A_3 : A_1^3)$   &  0   & edge-type self-linking with \\
   &   &     &   a Y-junction point \\  [1ex]

xvii)  &  $(A_3 : A_3)$   &  0   & edge-type self-linking with \\
  &   &     &   an edge point \\  [1ex]
\label{Table1} 
\end{longtable}

\section{Simplifying the Blum Linking Structure for Medical Imaging}
\label{S:sec7}
	For modeling configurations in 2D and 3D medical images, the 
strategy we propose builds upon that used in modeling single objects 
using M-reps and S-reps as developed by Pizer and coworkers at UNC 
MIDAG.  The Blum medial axis is replaced by a simpler skeletal structure 
where the skeletal set (also called the medial set) is replaced by a 
surface with boundary diffeomorphic to a $2$-dimensional disk, which is 
modeled by a rectangular grid with a multi-valued vector field at the 
vertices of the grid.  Then, the grid together with vectors is the template 
which is fitted to a specific object in each of a training set of medical 
images.  The size of the grid is experimentally chosen based on the object.  
The fitting involves an optimization process based on several criteria 
involving nonsingularity of the structure, regularity of the interpolated 
vector fields, nonsingularity of the constructed object boundary, and 
closeness of fit to an expert segmentation of the object. These criteria 
use in several ways the underlying mathematics of skeletal structures to 
numerically measure the criterion and contribute to the interpolation 
technique. \par  
The fitting of the templates provides a discrete set of data for each 
object in the training set; but the data are points on a product manifold.  A 
form of nonlinear PCA is applied to identify the most significant feature 
directions.  Then an iterative process is applied to further refine the fits 
and ultimately to yield a fitting process which is applied to new images 
using the preceding criteria and the statistical priors.  This method 
provides extremely high quality fit yielding high segmentation accuracy 
for a number of different cases that have been studied, and led to 
commercial software for imaging the male pelvic region.     \par
In several cases mentioned earlier where several objects in the same 
image are considered, additional user chosen positional information is 
attached which improves the positioning of the templates to the 
individual objects.  What the work here provides for a configuration of 
objects is an analysis of how the entire collection can be simultaneously 
modeled beginning with a Blum linking structure.  Because of several weak 
features and disadvantages of the Blum medial axis, this linking structure 
would be replaced by a more general but simplified skeletal linking 
structure.  This linking structure would still restrict to give an S-rep 
representation for each object.  By using skeletal sets without singular 
points (except the edge points), the number of linking types are reduced 
from $17$ to $9$.  Moreover, if the external linking axis can be simplified 
by, for example, avoiding self-linking, then this can be further reduce to 
$7$ types in Table \ref{Table1}:  i), iii), viii), x), xii), xiii), and xiv).  
These concern the singular structure of the external linking axis and the 
relation of the edges of the skeletal sets with each other and the singular 
points of the linking axis.  The added data for a fitted linking structure, 
beyond that for the S-rep data for each object, would include the values of 
the linking functions at the vertex points, and the points on the skeletal 
sets corresponding to the linking types.  The linking function values are 
easily included in the data, while for the linking types more investigation 
is needed to effectively include these so the statistical analysis includes 
all information.  \par 
Right now this has begun on a region in the neck which includes the three 
objects illustrated in Figure \ref{fig.1b} b).  There is still needed the 
detailed fitting for a training set with experimental decisions made on 
the size of the grids for the S-reps to also incorporate the extra data.  
This will involve the use of additional mathematical structure built upon 
that for skeletal structures, which is explained in the related paper 
\cite{DG2}.  Then, the process will proceed to arrive at a coherent 
template for the entire configuration and a specific optimization 
procedure for the fitting.

\end{document}